\documentclass{article}
\pdfoutput=1

\usepackage[preprint]{corl_2026} %
\usepackage{multirow} 
\usepackage[disable]{todonotes}
\usepackage{float}
\usepackage{tikz}
\definecolor{promptbg}{gray}{0.96}   %
\usepackage{listings}
\lstnewenvironment{GrayVerbatim}
  {\lstset{
    basicstyle=\footnotesize\ttfamily,
    breaklines=true,
    breakatwhitespace=true,
    backgroundcolor=\color{promptbg},
    frame=none,
    breakindent=0pt, 
    xleftmargin=4pt,
    xrightmargin=4pt,
    columns=fullflexible,
    keepspaces=true,
    inputencoding=utf8,
    extendedchars=true,
    literate={→}{{\textrightarrow}}1 {—}{{---}}1
  }}
  {}
  
\usepackage{float}
\long\def\xxnote#1#2#3{}
\long\def\rawnote#1#2{}

\ifx\hidenotes\undefined
  \AtBeginDocument{\setlength{\marginparwidth}{1.25\marginparwidth}}
  \makeatletter
  \long\def\xxnote#1#2#3{%
    \ifx\@captype\@undefined
      \ifvmode
      \else
      \fi
    \else
      \textcolor{#2}{\textbf{[#1:} #3\textbf{]}}%
    \fi
  }
  \long\def\rawnote#1#2{%
    \ifx\@captype\@undefined
      \ifvmode
      \else
      \fi
    \else
      \textcolor{#1}{\textbf{[}#2\textbf{]}}%
    \fi
  }
  \makeatother
\fi

\newcommand{\ApproachName}{$\texttt{MessyMem}$}

\definecolor{Queryable}{HTML}{A571C8}   %
\definecolor{Updateable}{HTML}{45B5A5}  %
\definecolor{Finegrained}{HTML}{F08A6E} %
\newcommand{\queryableC}[1]{\textcolor{Queryable}{\textbf{#1}}}
\newcommand{\updateableC}[1]{\textcolor{Updateable}{\textbf{#1}}}
\newcommand{\finegrainedC}[1]{\textcolor{Finegrained}{\textbf{#1}}}

\newcommand{\memtag}[2]{%
  \raisebox{0.3ex}{%
    \tikz[baseline=(X.base)]
      \node[
        draw=#1,
        text=#1,
        line width=0.4pt,
        rounded corners=2pt,
        inner xsep=4pt,
        inner ysep=1pt
      ] (X) {\scriptsize\sffamily #2};%
  }%
}

\newcommand{\GlobalTag}{\memtag{Queryable}{Globally queryable}}
\newcommand{\InteractTag}{\memtag{Updateable}{Updateable from interactions}}
\newcommand{\FineTag}{\memtag{Finegrained}{Fine-grained}}

\usepackage{booktabs}

\title{\ApproachName: Learning-from-Doing Memory for \\Mobile Manipulation}

\author{
\textbf{Anuva Banwasi} \quad
\textbf{William Muckelroy III} \quad
\textbf{Priya Sundaresan} \quad
\textbf{Linfeng Zhao}
\\
\textbf{Jeannette Bohg} \quad
\textbf{Cherie Ho}
\\[0.5em]
Stanford University
\\[0.2em]
\href{https://messymem.github.io}{\texttt{https://messymem.github.io}}
}

\begin{document}
\maketitle
\raggedbottom
\vspace{-3.0em}
\begin{figure}[H]
    \centering
\includegraphics[width=1.00\textwidth]{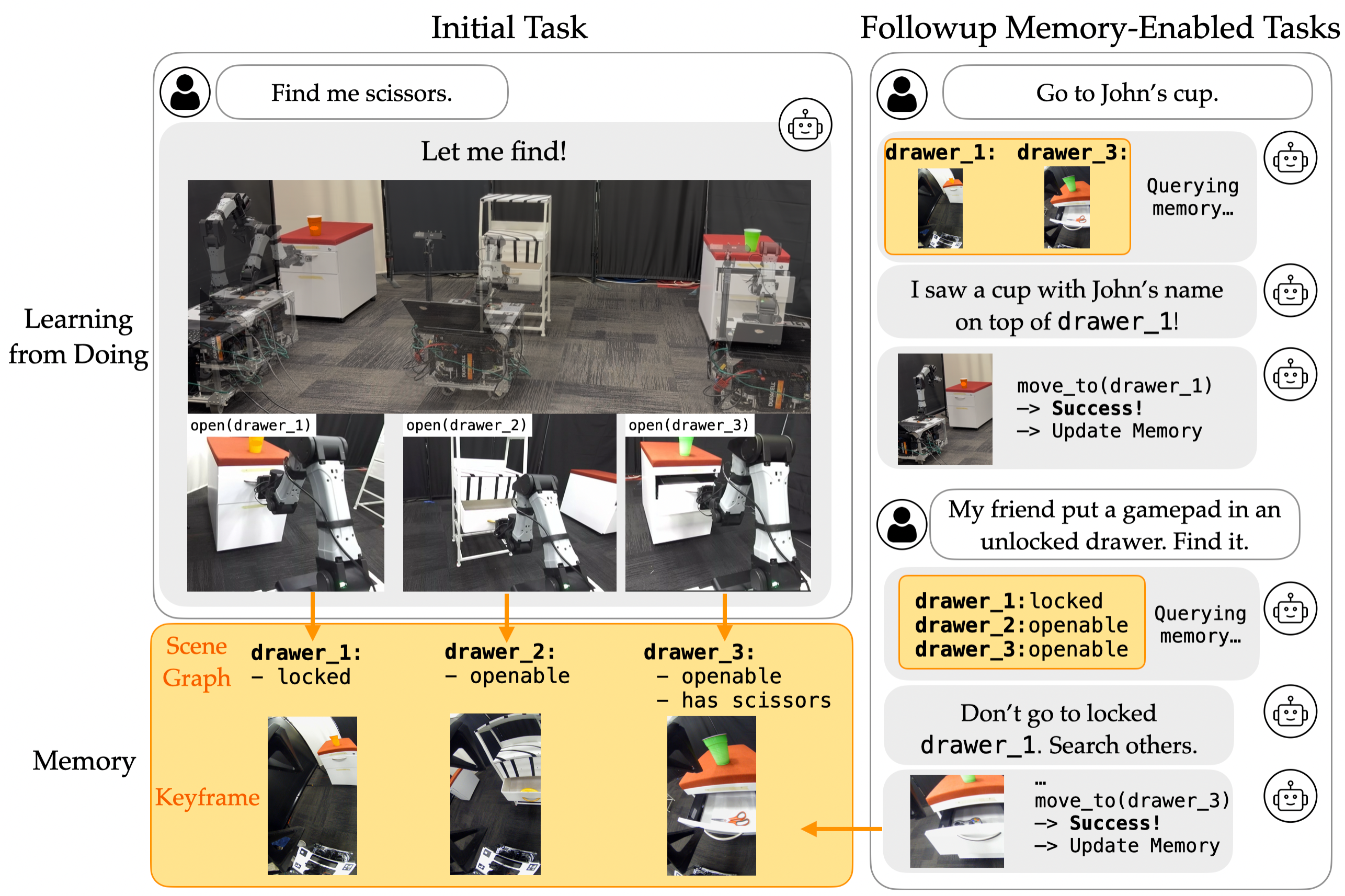}
    \caption{
    \small
    \textbf{\ApproachName{}} is a persistent mobile-manipulation memory system maintaining a 3D scene graph with interaction-derived properties and linked keyframes. \textbf{(Left)} The robot builds memory while exploring to \textit{Find me scissors}. \textbf{(Right)} On later open-vocabulary tasks, it reuses memory rather than re-exploring, retrieving fine-grained visual evidence to locate \textit{John's cup} and interaction-derived properties to avoid the locked drawer.}
    \label{fig:pull_figure}
\end{figure}

\begin{abstract}
Mobile manipulators deployed across many rooms and visits should improve with experience: after discovering that a cabinet is locked or finding an object in a drawer, the robot should reuse that knowledge rather than start each task from scratch. Yet today's robots often treat each task as new: compact scene representations omit interaction-derived knowledge, raw video histories are difficult to query, and VLM planners reason at inference time without persistently updating what the robot knows. We present \ApproachName{}, a persistent memory system that enables mobile manipulators to learn from experience and reuse that knowledge across future tasks. It maintains a spatially grounded 3D scene graph of objects and locations, augments it with properties and outcomes learned through interaction, and links visual observations for fine-grained recall. We evaluate \ApproachName{} in simulation and on a real mobile manipulator. In a continuous 25-task simulation spanning over 3 hours, \ApproachName{} achieves 80.0\% task progress, outperforming the strongest ablation by 14.8 percentage points and the strongest external baseline by 28.9 points, while retrieving task-relevant evidence from thousands of stored keyframes and over an hour into the past.
\end{abstract}

\keywords{Interaction-Grounded Memory, Mobile Manipulation, Scene Graph} 

\section{Introduction}
Imagine asking a household robot to bring a child their favorite mug. Without any extra context, the robot should be able to recognize it from a dinosaur sticker on the handle, a detail it noticed during yesterday's cleanup. Later, when asked to make coffee, it should head to the cabinet where the coffee grounds are kept, rather than scanning the whole kitchen. When searching for sugar to go with the coffee, it should remember which cabinets are locked, and only try the unlocked ones. These behaviors share a property current robots lack: a persistent memory of what observation and physical interaction have revealed, anchored across the space the robot works in over a long time. Without such memory, every task starts from scratch, such as re-opening every cabinet.

These examples point to three requirements for memory in mobile manipulation:\\
(1) \queryableC{Globally queryable \& spatially grounded.} Memory must support efficient queries across large environments (rooms, floors, buildings) and long deployments (across sessions), in a form compact enough to actually search rather than a sliding window that forgets or an unindexed video log that cannot be queried at scale. Each query must return not only \emph{what} the robot has seen but \emph{where} to go to find it, so the planner can pick a visit order and navigate directly instead of re-exploring. \\
(2) \updateableC{Updatable from interactions.} Many object properties (e.g., weight, emptiness, locked-vs-unlocked, articulation) cannot be inferred from passive observation and must be learned through physical interaction. Memory must learn and save what each interaction reveals.\\
(3) \finegrainedC{Fine-grained.} For new tasks, text-based summaries alone are often insufficient; the robot must be able to retrieve fine-grained visual context on demand (e.g., the exact shelf where coffee was last seen, the label on a particular container).

The concept of memory has a rich history in robotics, but has yet to deliver on the three above properties simultaneously. 3D scene representations such as scene graphs~\citep{hughes2022hydra, rosinol2021kimera, armenio20193dscenegraphs, mccormac2017semanticfusion, gu2023conceptgraphs} capture environment structure from passive observation, but miss what a robot learns from physical interaction. Raw video frames preserve fine-grained detail but are expensive to search; keyframe-selection and video-memory methods~\citep{sridhar2025memer, mark2026bpp, torne2026mem} provide more compact visual histories, but generally lack the persistent spatial grounding needed for mobile manipulation. Recent scene-graph and keyframe systems~\citep{yin2024sgnav, werby2025keysg, shan2025graph2nav} primarily target navigation. Closest to our setting, CuriousBot~\citep{wang2025curiousbot} and RoboEXP~\citep{jiang2024roboexp} build 3D scene graphs through interactive exploration, but do not pair properties learned through interaction with fine-grained visual evidence that can be retrieved for future tasks.

\textbf{Our key contribution is \ApproachName{}, a memory system that lets mobile manipulators carry knowledge across tasks, large spaces, and long timescales instead of re-exploring.} It combines three aforementioned memory properties that support flexible queries for diverse future tasks (Fig.~\ref{fig:pull_figure}).
\ApproachName~builds a compact 3D scene graph whose nodes are observed objects associated with their world-frame positions, keeping memory \queryableC{globally queryable and spatially grounded}. After an action, an interaction analyzer infers object properties that are only revealed through the robot's actions and attaches a summary to the corresponding node, keeping memory \updateableC{updatable from interactions}. Finally, it saves keyframes and links them to scene-graph nodes, enabling \finegrainedC{fine-grained} retrieval of image detail. 
By indexing both interaction-derived properties and keyframes through the scene graph, \ApproachName{} can retrieve a compact, task-relevant subset of past experience at plan time rather than searching through raw hours-long visual histories.

We evaluate \ApproachName{} in simulation and on a real TidyBot++ platform, including 50-trial controlled evaluations and a 25-task sequence spanning over 3 hours of continuous execution. Across these experiments, \ApproachName{} consistently outperforms its ablations and external memory baselines. On the long-horizon evaluation, it reaches 80.0\% task progress, 14.8 percentage points above the strongest ablation and 28.9 points above the strongest external baseline.

\section{Related Work}
\textbf{Queryable spatial memory.}
Metric-semantic maps and 3D scene graphs provide a natural representation for globally queryable and spatially grounded robot memory, from object-centric environment graphs~\citep{hughes2022hydra, rosinol2021kimera, armenio20193dscenegraphs, mccormac2017semanticfusion} to open-vocabulary maps that support language queries over objects and places~\citep{shafiullah2022clipfields, huang2022vlmaps, jatavallabhula2023conceptfusion, kerr2023lerf, peng2022openscene, gu2023conceptgraphs}.
Such maps can ground large-scale LLM planning~\citep{rana2023sayplan}, but are typically constructed from passive observations of geometry, location, and semantics. \ApproachName{} augments this structure with action-revealed information, including object properties such as locked, empty, or heavy, together with execution outcomes and failure reasons.

\textbf{Fine-grained visual memory.}
Embodied agents also benefit from retaining image, video, or keyframe histories rather than conditioning only on the current observation~\citep{savva2019habitat, parisotto2018neuralmap, parisotto2020stabilizing, baker2022video}.
Recent robot-memory systems build multimodal environment memories for embodied interactive agents~\citep{liu2024meia}, use 3D scene-memory snapshots for exploration and reasoning~\citep{yang20243dmem}, retrieve experience keyframes~\citep{sridhar2025memer}, select key history frames for imitation~\citep{mark2026bpp}, combine short-horizon video with long-horizon text memory~\citep{torne2026mem}, or build non-parametric embodied memories for retrieval and generation~\citep{xie2024embodiedrag}. These approaches preserve rich visual evidence, but generally do not couple it with persistent spatial structure and interaction-derived state. \ApproachName{} instead links keyframes to persistent scene-graph entities, keeping fine-grained visual evidence spatially grounded for future planning.

\textbf{Interaction-grounded memory and planning.}
Planning under partial observability treats hidden state as a belief-update problem, from POMDPs and decision making under uncertainty~\citep{kaelbling1998planning, kochenderfer2015decision, kochenderfer2022algorithms} to robotic task-and-motion planning in belief space~\citep{kaelbling2012unifying, kaelbling2013integrated, garrett2020online}.
Recent work also uses foundation models as uncertainty estimators for belief-space planning~\citep{zhao2025seeing}.
Interactive-perception systems infer articulation or actionability from physical trials~\citep{mo2021where2act, jiang2022ditto}. 

RoboEXP and CuriousBot are closest to our setting in using interaction to build action-conditioned or relational scene graphs~\citep{jiang2024roboexp, wang2025curiousbot}. These systems primarily enrich scene representations during exploration, and RoboEXP uses a predefined relational schema (e.g., \texttt{inside}, \texttt{on}). In contrast, \ApproachName{} updates persistent memory during task execution with open-vocabulary properties and outcomes that the planner can reuse across later tasks. We directly compare against RoboEXP under a shared execution stack.

\textbf{Planner-facing memory.}
Foundation-model robot planners reason flexibly from language and vision, including affordance-grounded planning~\citep{ahn2022saycan}, embodied multimodal planning~\citep{driess2023palm}, vision-language-action policies~\citep{brohan2023rt2}, long-horizon language-to-motion planning~\citep{lin2023text2motion}, and 3D value-map composition~\citep{huang2023voxposer}.
World-model and latent-memory methods learn implicit state for control~\citep{hafner2020dreamer, parisotto2018neuralmap}.
These systems provide strong inference-time reasoning, but do not expose an editable object-level memory that persists across tasks and combines structured state with retrievable visual evidence.

\noindent
Taken together, prior work provides complementary pieces of spatial structure, visual recall, interaction-derived state, and foundation-model planning. \ApproachName{} combines these in a persistent memory interface that can be accumulated and reused across tasks and long time horizons.

\section{Method}\label{sec:method}
\begin{figure}[t]
  \centering
  \includegraphics[width=\linewidth]{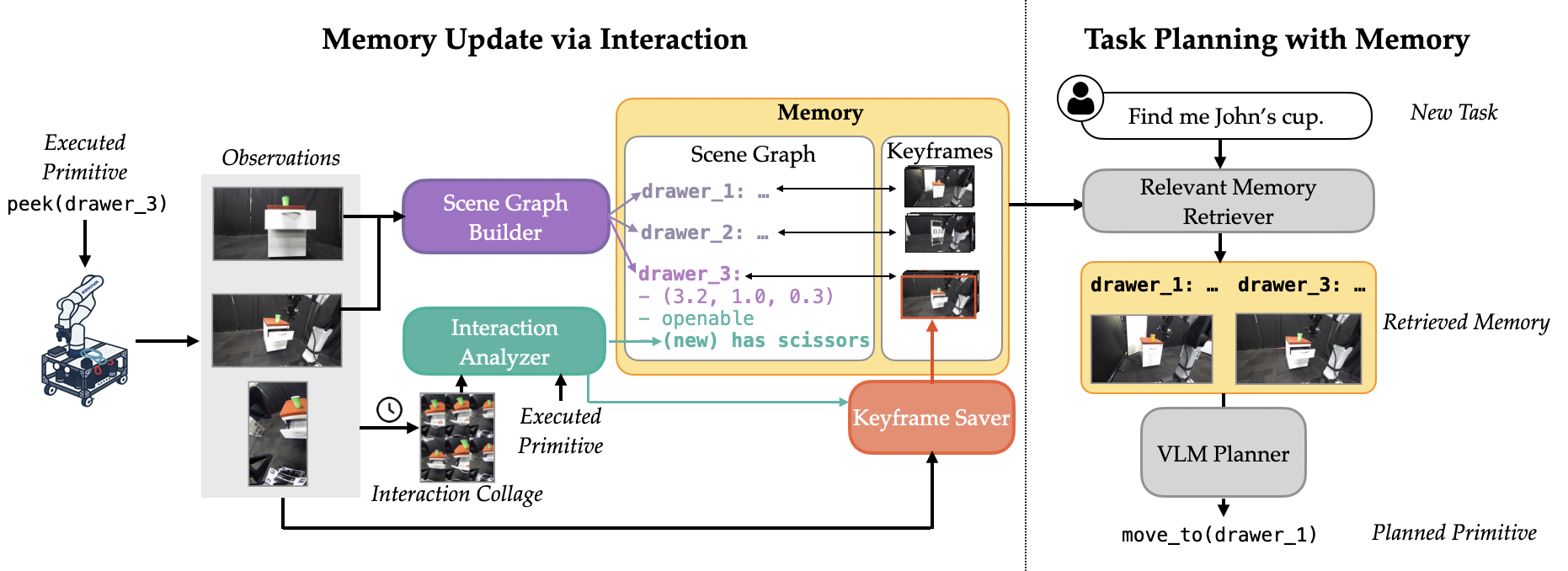}
  \vspace{-10px}
  \caption{
  \small
  \textbf{MessyMem approach overview.}
  \textbf{(Left)} A \textbf{scene graph builder} incrementally maintains a persistent 3D graph of observed objects and their locations, while a \textbf{keyframe saver} links selected visual observations to the corresponding scene-graph entries. 
  After manipulation, a VLM-based \textbf{interaction analyzer} reasons over the interaction collage and writes action-revealed properties, contents, and execution outcomes back to the scene graph.
  \textbf{(Right)} For each planning step, a \textbf{memory retriever} uses the task and scene graph to surface relevant structured properties and linked keyframes. 
  A \textbf{VLM planner} combines this retrieved memory with the current observation to select the next primitive; execution then produces new evidence that updates memory for subsequent decisions.
  }
  \vspace{-10px}
  \label{fig:method}
\end{figure}

We propose \ApproachName{}, a persistent memory system for mobile manipulation (Fig.~\ref{fig:method}). 
The robot incrementally builds memory from RGB-D observations and manipulation outcomes. Execution proceeds in a closed loop: retrieve task-relevant memory, plan and execute a primitive, analyze its outcome, update memory, and replan against the updated state.

\subsection{\queryableC{Globally Queryable \& Spatially Grounded} Text Memory: 3D Scene Graph}
\label{sec:sg}

We maintain a persistent 3D scene graph keyed by object instances. This compact structure serves as the index to which interaction-derived properties (§\ref{sec:analyzer}) and keyframes (§\ref{sec:kf}) are attached. A \textbf{scene graph builder} constructs the graph incrementally from incoming RGB-D observations. Each image is passed to SAM3~\cite{carion2025sam3segmentconcepts} to segment objects from a predefined set of category prompts; each mask is deprojected to 3D using the depth image, and same-label detections are clustered and associated with existing entries (details in Appendix). Each entry stores an \texttt{object\_id} (e.g., \texttt{drawer\_3}), a 3D centroid for spatial grounding, and a property dictionary maintained by the interaction analyzer.

Previously observed objects remain in the graph after leaving the current camera view. Interaction-derived properties and visual keyframes are associated with the same persistent entries, allowing remembered state and visual evidence to remain grounded to locations that the planner can revisit.

\subsection{\updateableC{Interaction-Updated} Text Memory: VLM Interaction Analyzer}
\label{sec:analyzer}

After each manipulation primitive, a \textbf{VLM interaction analyzer} converts the corresponding visual interaction window into structured updates to the targeted scene-graph entry. This captures information that passive observation alone may not reveal, including object properties such as whether a cabinet is \texttt{locked} or \texttt{openable}, newly discovered fixture contents, and execution outcomes such as success, failure, or failure reason.

The system collects the corresponding action window, forms a multi-frame collage, and queries the VLM-based analyzer. Reasoning over the interaction window rather than only a single before/after observation exposes how the action unfolds: for example, whether the gripper misses a handle, slips during a pull, or successfully grasps while the cabinet itself remains stationary. The analyzer parses this evidence into structured fields and writes them to the relevant scene-graph entry. It updates a fixed set of broadly useful properties (e.g., \texttt{pickable}, \texttt{fullness}, \texttt{openable}), can add open-vocabulary properties when useful (e.g., \texttt{material}, \texttt{has\_handles}), records execution outcomes and failure reasons, and refreshes fixture \texttt{contents} from newly observed evidence.

Multi-frame context allows the analyzer to distinguish visually similar end states arising from different causes, such as a missed grasp versus a successful grasp on a locked cabinet (Appendix Fig.~\ref{fig:interaction-analyzer-fig}). For example, after \texttt{peek(drawer\_3)}, the analyzer may write \texttt{contents: scissors} to the \texttt{drawer\_3} node; after an unsuccessful opening attempt, it can distinguish a missed grasp from evidence that the drawer is locked.

New evidence updates the corresponding current scene-graph state. If later observations reveal different contents or object state, the newer value overwrites the stale structured property, allowing outdated or incorrect memory to be corrected. Keyframes, in contrast, retain the underlying visual history rather than being overwritten.

\subsection{\finegrainedC{Fine-Grained} Visual Memory: Linked Keyframes}
\label{sec:kf}

The \textbf{keyframe saver} maintains a bank of selected RGB observations linked to the persistent scene-graph entries they describe. Keyframes are saved from both interactions and scene updates. Around each manipulation action, the system stores the frame before the action, the frame after the action, and the frame the interaction analyzer identifies as most informative; these frames are linked to the scene-graph entry targeted by the action. Outside manipulation, frames are also saved when the scene graph registers a new object or updates an existing entry, and are linked to the visible entries. Near-duplicate observations are removed to reduce redundant growth (details in Appendix).

Each keyframe stores a full-resolution RGB image and a downsampled thumbnail together with metadata such as its timestamp, originating action, and linked scene-graph entries. Interaction-derived keyframes can additionally receive short free-text labels from the analyzer (e.g., \texttt{spoiled bananas}, \texttt{open drawer interior}), providing retrieval cues for visual contents that may not correspond directly to a scene-graph object label.

Together, the scene graph maintains compact object-centric state and spatial grounding, while linked keyframes retain appearance, text, clutter, and other instance-level details from past observations.

\subsection{Memory Retrieval and VLM Task Planning}
\label{sec:retrieval}

At each planning step, the \textbf{memory retriever} first surfaces a compact, task-relevant subset of the accumulated memory. The \textbf{VLM planner} then combines this evidence with the current observation and scene graph to select the next primitive.

Given the task description and current scene-graph state, the retriever builds a candidate pool from three complementary sources: (1) keyframes linked to scene-graph entries, (2) keyframes whose analyzer-written labels match content in the task, and (3) recent interaction keyframes that provide short-term context. Candidates from these sources are merged and deduplicated, with per-source caps preventing recently observed entities from overwhelming the retrieval pool.

Retrieval then proceeds in two stages. In the first, text-only stage, a VLM receives a compact catalog describing each candidate by its linked scene-graph entries, labels, originating action, trigger, and other metadata, together with the task and a compact scene-graph summary. It selects a shortlist of frames likely to help solve the task. In the second, multimodal stage, the VLM inspects thumbnails of the shortlisted frames and selects a small set of clear, non-redundant observations for the planner. This staged design uses structured memory to first narrow the candidate set, then applies visual reasoning only to the most relevant frames. Appendix Fig.~\ref{fig:keyframes-retrieval} illustrates an example in which retrieved keyframes preserve an instance-specific cue, the name printed on a cup, that is absent from the structured scene graph.

The VLM planner receives the retrieved full-resolution frames together with short descriptions of why they were selected, the full scene graph as JSON, the current camera observation, the task description, and the available motion primitives. It selects the next primitive and scene-graph target; after execution, the interaction analyzer updates the relevant memory entries, and the next planning step retrieves again against the newly updated state.

\section{Experimental Setup}
We evaluate \ApproachName{} on memory-intensive scenarios in simulation and the real world.

\paragraph{Evaluation Protocol.}
\textit{Initialization}: Each trial begins with a predefined trajectory that observes fixture locations (e.g., cabinets, drawers, counters) without revealing their contents; contents become observable only after interaction.
\textit{Task sequence}: Tasks are issued sequentially after the previous task completes or fails, requiring information acquired earlier to persist across decisions.
\textit{Primitives}: All methods use the same navigation, fixture-interaction, and manipulation primitives, with embodiment-specific details in the Appendix.
\textit{Metrics}: We report task progress~\cite{torne2026mem,sridhar2025memer}, defined by task-specific success criteria, together with the number of actions. For the two short simulation scenarios, we run 50 trials per method and report mean task progress with 95\% confidence intervals.

\paragraph{Compared Methods.}
All methods share the same planner, perception stack, and action primitives, isolating the effect of the memory representation.

\begin{list}{$\bullet$}{%
  \setlength{\leftmargin}{1.35em}%
  \setlength{\labelwidth}{0.8em}%
  \setlength{\labelsep}{0.45em}%
  \setlength{\itemsep}{0pt}%
  \setlength{\parsep}{0pt}%
  \setlength{\topsep}{0.5ex}%
}

\item
\textbf{\ApproachName{} (SG+IA+KF):}
Our full method, combining a persistent scene graph (SG), interaction-analyzer updates (IA), and linked keyframe memory (KF).

\item
\textbf{SG+IA:}
\ApproachName{} without keyframe memory. This tests whether spatially grounded structured memory and interaction-derived properties are sufficient without access to fine-grained visual history.

\item
\textbf{SG+KF:}
\ApproachName{} without the interaction analyzer. The method retains object positions and linked keyframes but does not store properties or outcomes inferred from interactions.

\item
\textbf{SG:}
A scene graph containing object labels and positions, without interaction-derived updates or keyframe memory. This tests the limit of a globally queryable spatial representation alone.

\item
\textbf{RoboEXP~\cite{jiang2024roboexp}:}
We adapt RoboEXP's Action-Conditioned Scene Graph (ACSG) memory to our execution stack. Its predefined relational representation stores structured spatial and action-conditioned information, while the planner, perception, and primitives remain shared with the other methods.

\item
\textbf{MemER-style~\cite{sridhar2025memer}:}
We implement MemER's episodic visual-memory paradigm, in which a VLM selects salient observations that are clustered temporally into representative keyframes. The planner reasons from these episodic frames without \ApproachName{}'s persistent scene-graph indexing or interaction-derived state.

\item
\textbf{MemER+Fixtures:}
To separate episodic-memory limitations from basic mobile-navigation grounding, we additionally provide MemER-style memory with the identities and positions of visible and out-of-view fixtures.

\end{list}

\subsection{Simulation Scenarios}

\textbf{Environment.}
We instantiate our scenarios in the MuJoCo~\cite{todorov2012mujoco}-based RoboCasa365 simulator~\cite{robocasa365} for household mobile manipulation. Full scenario definitions and success criteria in Appendix.

\textbf{Locked-and-Unlocked Cabinets}:
\GlobalTag~\InteractTag~
Among three cabinets, one is locked but its status is unknown before interaction. The robot must first \textit{look for a banana} and then \textit{look for ketchup}. Solving the second task requires retaining both cabinet contents and the previously discovered locked status.

\textbf{Clutter-Aware Pick}:
\GlobalTag~\InteractTag~\FineTag~
Across two cabinets, the robot first \textit{finds the mustard} and is then asked to \textit{pick up the ketchup}. Ketchup appears in both cabinets, but only one instance is readily accessible; the other is embedded in surrounding clutter. Correct selection requires recalling fine-grained visual context beyond object identity or presence.

\textbf{25-task Long Horizon}:
\GlobalTag~\InteractTag~\FineTag~
We evaluate a continuous sequence of 25 household tasks in a 10-cabinet kitchen containing more than 60 objects and fixtures. The sequence spans four task families: \textit{finding}, \textit{matching}, \textit{manipulation}, and \textit{memory queries}, and executes for over 3 hours without resetting memory. Tasks require recalling observations from much earlier in the sequence, distinguishing visually similar object instances, reasoning about where household objects belong, and reusing information acquired during prior manipulation. Examples include matching a cup to one observed earlier, identifying a reusable bottle among similar variants, recalling which objects were stored together, and finding a mug placed away earlier.

\subsection{Real-World Scenarios}

\textbf{Hardware.}
We use the TidyBot++ platform~\cite{wu2024tidybotpp} with a 6-DoF YAM arm and holonomic mobile base. Two ZED~2 RGB-D cameras support scene-graph construction and 3D object localization: a front-facing base camera and a pole-mounted camera that provides a higher vantage point and observations to the VLM planner. A wrist-mounted Arducam fisheye provides additional views for keyframe memory, while wall-mounted fiducial markers support localization.

\textbf{Office Drawer Search}:
~\GlobalTag~\InteractTag~\FineTag~
Across three office drawers, one of which is locked, the robot sequentially searches for scissors, identifies John's cup, and searches for a newly placed game controller in an unlocked drawer. The sequence requires persistent spatial grounding, reuse of the discovered lock status, and fine-grained visual recall to distinguish objects using details that may not be captured in the scene graph.

\textbf{Sock Pairing}:
~\GlobalTag~\InteractTag~\FineTag~
With two sock-filled drawers and one unpaired sock on a chair, the robot first inspects both drawers and then places the loose sock into the drawer containing its matching pair. Since candidate socks can share coarse attributes such as color, success depends on recalling fine-grained appearance, including pattern, while retaining the spatial association between the relevant visual memory and its drawer.

\section{Experimental Results}
\label{sec:result}

\begin{table*}[t]
\footnotesize
\centering
\setlength{\tabcolsep}{5pt}
\begin{tabular}{lccc}
\toprule
\textbf{Method}
& \textbf{Cluttered Pick}
& \textbf{Locked / Unlocked}
& \textbf{25-task Long Horizon} \\
& Task Prog. $\uparrow$
& Task Prog. $\uparrow$
& Task Prog. $\uparrow$ \\
\midrule

\textbf{\ApproachName{} (Ours)}
& \textbf{84.0 [77.0, 90.0]}
& \textbf{99.0 [97.0, 100.0]}
& \textbf{80.0 [77.4, 82.6]} \\

SG + IA
& 58.0 [53.0, 63.0]
& 88.0 [82.0, 94.0]
& 65.2 [63.5, 67.0] \\

RoboEXP~\cite{jiang2024roboexp}
& 53.0 [48.0, 58.0]
& 70.0 [63.0, 76.0]
& 51.1 [49.1, 53.1] \\

SG + KF
& 51.0 [44.0, 58.0]
& 43.0 [35.0, 51.0]
& 47.9 [45.5, 50.4] \\

SG
& 49.0 [45.0, 53.0]
& 50.0 [50.0, 50.0]
& 49.1 [47.4, 51.0] \\

MemER-style~\cite{sridhar2025memer}
& 0.0 [0.0, 0.0]
& 0.0 [0.0, 0.0]
& 1.8 [1.0, 2.5] \\

MemER + Fixtures
& 8.0 [3.0, 13.0]
& 27.0 [16.0, 39.0]
& 0.7 [0.3, 1.2] \\

\bottomrule
\end{tabular}
\caption{
\textbf{\ApproachName{} consistently outperforms its ablations and external memory baselines.}
Mean task progress [95\% CI] over 50 trials per method. The short scenarios isolate fine-grained visual and interaction-derived memory, while Long Horizon evaluates all memory properties jointly over 25 consecutive tasks.
}
\label{tab:sim_results}
\end{table*}

\subsection{Simulation Results}

Table~\ref{tab:sim_results} shows that \ApproachName{} achieves the highest task progress across all three simulated evaluations.
No individual memory component suffices across scenarios: interaction-derived state is particularly useful when the robot must reuse properties learned through action, while linked keyframes preserve visual information that is difficult to capture in a compact structured representation.
Combining both with a persistent spatial scene graph yields the strongest performance across tasks.

\paragraph{Fine-grained visual memory enables instance-level reasoning.}
On \textit{Cluttered Pick}, \ApproachName{} reaches \textbf{84\%} task progress, compared with \textbf{58\%} for SG+IA and \textbf{51\%} for SG+KF. Without keyframes, the robot can remember where ketchup was observed but cannot recover the detailed visual arrangement needed to select the less obstructed instance. The full method combines structured grounding with linked keyframes that recover this task-relevant visual evidence.

\paragraph{Interaction-derived memory captures action outcomes.}
On \textit{Locked / Unlocked}, \ApproachName{} reaches \textbf{99\%} progress and SG+IA \textbf{88\%}, compared with \textbf{43\%} for SG+KF. A saved image alone may not distinguish a locked cabinet from a failed grasp. The Interaction Analyzer reasons over a higher-frequency collage spanning the manipulation, allowing it to determine success or failure, infer properties revealed by the action, and store the resulting state in the scene graph for later tasks.

\paragraph{Comparison with external memory systems.}
RoboEXP reaches \textbf{53\%}/\textbf{70\%} on Cluttered Pick and Locked / Unlocked, compared with \ApproachName{} at \textbf{84\%}/\textbf{99\%}. Its predefined Action-Conditioned Scene Graph captures spatial structure but not fine-grained appearance or richer interaction-derived state (e.g., a locked cabinet). \ApproachName{} instead stores action-revealed properties and outcomes while linking keyframes for instance-level visual recall.
 
We find that MemER-style memory struggles in mobile manipulation, where relevant fixtures leave view and keyframes alone lack persistent spatial grounding. Providing visible and out-of-view fixture identities and positions improves MemER+Fixtures from \textbf{0\%} to \textbf{8\%}/\textbf{27\%} on the two short scenarios but performance remains low.

\subsection{Long-Horizon Memory}

The 25-task Long Horizon evaluation tests how these differences compound over extended operation. Across more than 3 hours, \ApproachName{} achieves \textbf{80.0\%} task progress, outperforming the strongest ablation, SG+IA, by \textbf{14.8 percentage points} and the strongest external baseline, RoboEXP, by \textbf{28.9 points}. All six paired comparisons remain significant after Holm correction ($p_{\mathrm{Holm}}<10^{-4}$). The ablations show that neither structured nor visual memory alone is sufficient across a diverse task sequence. SG+IA reaches \textbf{65.2\%}: interaction-updated structured state supports many tasks but cannot recover visual details absent from the graph, limiting matching and instance-level recall. SG+KF (\textbf{47.9\%}) retains visual history but cannot persist what prior actions revealed, while SG alone (\textbf{49.1\%}) lacks both capabilities. \ApproachName{} combines them, using compact structured state when sufficient and retrieving linked visual evidence when finer detail is required.

RoboEXP reaches \textbf{51.1\%} on Long Horizon, exposing the limitation of predefined structured relations for instance-level distinctions. For example, remembering that a \texttt{cereal\_box} is inside a cabinet does not distinguish Froot Loops from a visually similar Trix box; \ApproachName{} can instead retrieve linked keyframes to recover the needed visual evidence. MemER-style and MemER+Fixtures reach only \textbf{1.8\%} and \textbf{0.7\%}, respectively. Over 3+ hours, MemER's recency bias makes older evidence increasingly difficult to surface, whereas \ApproachName{} retrieves task-relevant keyframes using the current goal and scene-graph associations.

\paragraph{Memory remains useful as it grows.}

\ApproachName{} accumulates a median of \textbf{9,669 keyframes} with a median retrieval latency of \textbf{8.6\,s}. The current goal and scene-graph associations narrow thousands of observations to a small candidate set before visual reranking, enabling a median lookback of \textbf{62.7 minutes}. Task-relevant experience remains accessible over hours of operation.

\subsection{Real-World Results}

The simulation findings transfer to the real robot: \ApproachName{} achieves \textbf{1.00} task progress in both scenarios, while each ablation exposes a distinct failure mode under real-world variability.

\begin{table*}[t]
\scriptsize
\centering
\setlength{\tabcolsep}{2.5pt}
\begin{tabular}{l|ccc|ccc|ccc|ccc}
\toprule
\textbf{Method}
& \multicolumn{3}{c|}{\textbf{1: ``Find Scissors''}}
& \multicolumn{3}{c|}{\textbf{2: ``Move to John's cup''}}
& \multicolumn{3}{c|}{\textbf{3: ``Find placed controller''}}
& \multicolumn{3}{c}{\textbf{Overall}} \\
& Task Prog.$\uparrow$ & Steps$\downarrow$ & SPL$\uparrow$
& Task Prog.$\uparrow$ & Steps$\downarrow$ & SPL$\uparrow$
& Task Prog.$\uparrow$ & Steps$\downarrow$ & SPL$\uparrow$
& Task Prog.$\uparrow$ & Steps$\downarrow$ & SPL$\uparrow$ \\
\midrule

\ApproachName{} (Ours)
& \textbf{1.00} & \textbf{6.4} & 0.85
& \textbf{1.00} & 1.2 & \textbf{0.90}
& \textbf{1.00} & 3.2 & \textbf{1.00}
& \textbf{1.00} & \textbf{10.8} & \textbf{0.92} \\

SG + IA
& \textbf{1.00} & 6.6 & \textbf{0.90}
& 0.60 & 1.4 & 0.60
& \textbf{1.00} & \textbf{3.0} & 0.96
& 0.87 & 11.0 & 0.82 \\

SG + KF
& 0.80 & 7.4 & 0.67
& 0.60 & 3.6 & 0.34
& 0.40 & 6.0 & 0.19
& 0.60 & 17.0 & 0.40 \\

SG
& 0.00 & 9.4 & 0.00
& 0.20 & \textbf{1.0} & 0.20
& \textbf{1.00} & 5.0 & 0.81
& 0.40 & 15.4 & 0.34 \\
\bottomrule
\end{tabular}
\caption{
\textbf{Per-task breakdown for \textit{Office Drawer Search}.}
Task progress, steps, and success-weighted path length (SPL) are reported for each sequential goal; overall steps are summed across the trial ($n=5$). SPL is computed per trial and averaged using human-annotated reference step counts ($L=6,1,4$ for Tasks 1--3), reflecting a reasonable human path, rather than a literal minimum.
}
\label{tab:office_drawer_search_results}
\end{table*}

\paragraph{Reusing interaction outcomes and visual evidence.}
In \textit{Office Drawer Search}, \ApproachName{} is the only method to complete all three sequential goals, reaching \textbf{1.00} overall progress in \textbf{10.8} steps on average while following closest to the reference path with an SPL of \textbf{0.92}; $n=5$ trials (Table~\ref{tab:office_drawer_search_results}). During drawer exploration, IA-enabled methods can reuse the discovered lock status rather than repeatedly attempting an infeasible interaction; \ApproachName{} completes the first goal in \textbf{6.4} steps, compared with 7.4 for SG+KF and 9.4 for SG. The second goal, ``Move to John's cup,'' additionally requires fine-grained visual information absent from the scene graph: \ApproachName{} reaches \textbf{1.00} progress, while SG+IA reaches \textbf{0.60}. Across the full sequence, methods lacking either interaction-updated state or fine-grained visual recall fail more often and require more actions.

\begin{figure}[t]
    \centering
    \includegraphics[width=\linewidth]{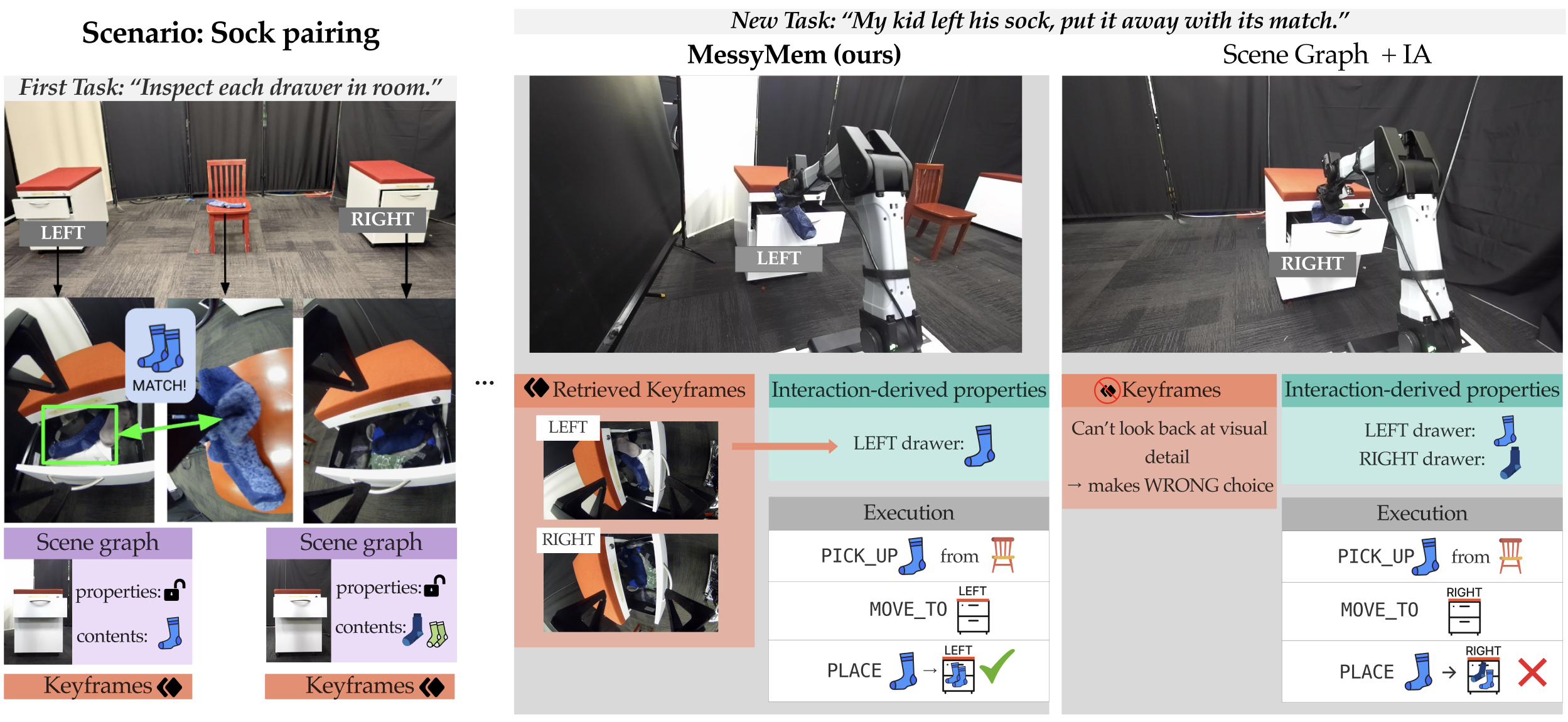}
    \caption{
    \small
    \textbf{\textit{Sock Matching} scenario.}
    During inspection, \ApproachName{} links keyframes of the left and right drawer interiors to its scene graph. When asked to match the loose sock, the VLM planner retrieves these fine-grained observations, compares them against the target sock, and returns to the corresponding drawer. Without keyframes, \textbf{SG+IA} cannot recall the needed visual detail and selects the wrong drawer.
    }
    \label{fig:real_sock_results}
\end{figure}

\paragraph{Fine-grained recall for visual matching.}
The \textit{Sock Pairing} scenario isolates fine-grained visual recall in a spatially grounded setting: after inspecting two drawers, the robot must match a loose sock to a previously observed instance that differs from other candidates primarily in appearance. \ApproachName{} retrieves the linked drawer keyframes and compares them against the target sock, yielding \textbf{1.00} task progress versus \textbf{0.67} for SG+IA and SG+KF and \textbf{0.50} for SG over $n=6$ trials (Fig.~\ref{fig:real_sock_results}).

\section{Conclusion}
\label{sec:conclusion}
We present \ApproachName{}, a persistent memory system that enables mobile manipulators to learn from interaction and reuse experience across tasks, large spaces, and long timescales. \ApproachName{} combines a spatially grounded 3D scene graph, interaction-derived object properties and outcomes, and linked keyframes that preserve fine-grained visual evidence, yielding memory that is simultaneously \queryableC{globally queryable}, \updateableC{updateable from interaction}, and \finegrainedC{fine-grained}. Across 50-trial simulation evaluations, a 25-task sequence spanning over 3 hours of continuous execution, and real-world robot experiments, \ApproachName{} consistently outperforms its ablations and external memory baselines. On the long-horizon evaluation, it reaches 80.0\% task progress, 14.8 percentage points above the strongest ablation and 28.9 points above the strongest external baseline, while retrieving useful evidence from thousands of stored keyframes and over an hour into the past. These results show that structured state, interaction-derived updates, and accessible visual history provide complementary capabilities for persistent robot memory. \ApproachName{} is a step toward robots that can accumulate, recall, and act on experience throughout extended operation in complex real-world environments.

\textbf{Limitations.} Our current implementation uses a predefined closed-set prompt for object detections. A natural extension is to incorporate open-vocabulary detectors so the memory can capture more diverse objects and attributes. Our interaction analyzer is also focused on robot-initiated actions; extending it to reason over human actions would broaden the framework to HRI settings where memory must be updated from both robot and human activity.
Future work could also extend the same memory-conditioned interface to richer policy primitives, including VLA-based policies.

\clearpage
\acknowledgments{This work is supported by the Toyota Research Institute. This work used Marlowe~\cite{marlowe2025}, Stanford University's GPU-based Computational Instrument, supported by Stanford HAI and Stanford Research Computing. Priya Sundaresan is supported by a NSF GRFP. Cherie Ho is supported by the Croucher Postdoctoral Fellowship. Thank you to Satvik Sharma and Francis Engelmann for helpful discussions. We thank Alberta Longhini, Tyler Lum, and Carlota Parés-Morlans for feedback on earlier drafts of the paper. The views and conclusions contained
herein are those of the authors and should not be interpreted as necessarily representing the official
policies, either expressed or implied, of the sponsors. }

\bibliography{example}  %

\begin{thebibliography}{49}
\providecommand{\natexlab}[1]{#1}
\providecommand{\url}[1]{\texttt{#1}}
\expandafter\ifx\csname urlstyle\endcsname\relax
  \providecommand{\doi}[1]{doi: #1}\else
  \providecommand{\doi}{doi: \begingroup \urlstyle{rm}\Url}\fi

\bibitem[Hughes et~al.(2022)Hughes, Chang, and Carlone]{hughes2022hydra}
N.~Hughes, Y.~Chang, and L.~Carlone.
\newblock Hydra: A real-time spatial perception system for {3D} scene graph construction and optimization.
\newblock In \emph{Robotics: Science and Systems (RSS)}, 2022.

\bibitem[Rosinol et~al.(2020)Rosinol, Abate, Chang, and Carlone]{rosinol2021kimera}
A.~Rosinol, M.~Abate, Y.~Chang, and L.~Carlone.
\newblock Kimera: an open-source library for real-time metric-semantic localization and mapping.
\newblock In \emph{IEEE International Conference on Robotics and Automation (ICRA)}, 2020.

\bibitem[Armeni et~al.(2019)Armeni, He, Gwak, Zamir, Fischer, Malik, and Savarese]{armenio20193dscenegraphs}
I.~Armeni, Z.-Y. He, J.~Gwak, A.~R. Zamir, M.~Fischer, J.~Malik, and S.~Savarese.
\newblock 3d scene graph: A structure for unified semantics, 3d space, and camera.
\newblock In \emph{Proceedings of the IEEE International Conference on Computer Vision}, pages 5664--5673, 2019.

\bibitem[McCormac et~al.(2017)McCormac, Handa, Davison, and Leutenegger]{mccormac2017semanticfusion}
J.~McCormac, A.~Handa, A.~J. Davison, and S.~Leutenegger.
\newblock Semanticfusion: Dense 3d semantic mapping with convolutional neural networks.
\newblock In \emph{IEEE International Conference on Robotics and Automation (ICRA)}, 2017.

\bibitem[Gu et~al.(2024)Gu, Kuwajerwala, Morin, Jatavallabhula, Sen, Agarwal, Rivera, Paul, Ellis, Chellappa, Gan, de~Melo, Tenenbaum, Torralba, Shkurti, and Paull]{gu2023conceptgraphs}
Q.~Gu, A.~Kuwajerwala, S.~Morin, K.~M. Jatavallabhula, B.~Sen, A.~Agarwal, C.~Rivera, W.~Paul, K.~Ellis, R.~Chellappa, C.~Gan, C.~M. de~Melo, J.~B. Tenenbaum, A.~Torralba, F.~Shkurti, and L.~Paull.
\newblock Conceptgraphs: Open-vocabulary 3d scene graphs for perception and planning.
\newblock In \emph{IEEE International Conference on Robotics and Automation (ICRA)}, pages 5021--5028, 2024.

\bibitem[Sridhar et~al.(2025)Sridhar, Pan, Sharma, and Finn]{sridhar2025memer}
A.~Sridhar, J.~Pan, S.~Sharma, and C.~Finn.
\newblock Memer: Scaling up memory for robot control via experience retrieval.
\newblock \emph{arXiv preprint arXiv:2510.20328}, 2025.

\bibitem[Mark et~al.(2026)Mark, Attarian, Fu, Dwibedi, Liang, Shah, and Kumar]{mark2026bpp}
M.~S. Mark, M.~Attarian, C.~Fu, D.~Dwibedi, J.~Liang, D.~Shah, and A.~Kumar.
\newblock Bpp: Long-context robot imitation learning by focusing on key history frames.
\newblock In \emph{Robotics: Science and Systems (RSS)}, 2026.

\bibitem[Torne et~al.(2026)Torne, Pertsch, Walke, Vedder, Nair, Ichter, Ren, Wang, Tang, Stachowicz, Dhabalia, Equi, Vuong, Springenberg, Levine, Finn, and Driess]{torne2026mem}
M.~Torne, K.~Pertsch, H.~Walke, K.~Vedder, S.~Nair, B.~Ichter, A.~Z. Ren, H.~Wang, J.~Tang, K.~Stachowicz, K.~Dhabalia, M.~Equi, Q.~Vuong, J.~T. Springenberg, S.~Levine, C.~Finn, and D.~Driess.
\newblock Mem: Multi-scale embodied memory for vision language action models.
\newblock \emph{arXiv preprint arXiv:2603.03596}, 2026.

\bibitem[Yin et~al.(2024)Yin, Xu, Wu, Zhou, and Lu]{yin2024sgnav}
H.~Yin, X.~Xu, Z.~Wu, J.~Zhou, and J.~Lu.
\newblock Sg-nav: Online 3d scene graph prompting for llm-based zero-shot object navigation.
\newblock In \emph{Advances in Neural Information Processing Systems (NeurIPS)}, 2024.

\bibitem[Werby et~al.(2025)Werby, Rotondi, Scaparro, and Arras]{werby2025keysg}
A.~Werby, D.~Rotondi, F.~Scaparro, and K.~O. Arras.
\newblock Keysg: Hierarchical keyframe-based 3d scene graphs.
\newblock \emph{arXiv preprint arXiv:2510.01049}, 2025.

\bibitem[Shan et~al.(2025)Shan, Rajvanshi, Mithun, and Chiu]{shan2025graph2nav}
T.~Shan, A.~Rajvanshi, N.~Mithun, and H.-P. Chiu.
\newblock Graph2nav: 3d object-relation graph generation to robot navigation.
\newblock In \emph{IEEE International Conference on Robotics and Automation (ICRA)}, pages 1646--1652, 2025.

\bibitem[Wang et~al.(2026)Wang, Fermoselle, Kelestemur, Wang, and Li]{wang2025curiousbot}
Y.~Wang, L.~Fermoselle, T.~Kelestemur, J.~Wang, and Y.~Li.
\newblock Curiousbot: Interactive mobile exploration via actionable 3d relational object graph.
\newblock \emph{IEEE Robotics and Automation Letters}, pages 4993--5000, 2026.

\bibitem[Jiang et~al.(2024)Jiang, Huang, Wu, Li, Garg, Nayyeri, Wang, and Li]{jiang2024roboexp}
H.~Jiang, B.~Huang, R.~Wu, Z.~Li, S.~Garg, H.~Nayyeri, S.~Wang, and Y.~Li.
\newblock Roboexp: Action-conditioned scene graph via interactive exploration for robotic manipulation.
\newblock In \emph{Conference on Robot Learning (CoRL)}, 2024.

\bibitem[Shafiullah et~al.(2023)Shafiullah, Paxton, Pinto, Chintala, and Szlam]{shafiullah2022clipfields}
N.~M.~M. Shafiullah, C.~Paxton, L.~Pinto, S.~Chintala, and A.~Szlam.
\newblock Clip-fields: Weakly supervised semantic fields for robotic memory.
\newblock In \emph{Robotics: Science and Systems (RSS)}, 2023.

\bibitem[Huang et~al.(2023)Huang, Mees, Zeng, and Burgard]{huang2022vlmaps}
C.~Huang, O.~Mees, A.~Zeng, and W.~Burgard.
\newblock Visual language maps for robot navigation.
\newblock In \emph{IEEE International Conference on Robotics and Automation (ICRA)}, pages 10608--10615, 2023.

\bibitem[Jatavallabhula et~al.(2023)Jatavallabhula, Kuwajerwala, Gu, Omama, Chen, Maalouf, Li, Iyer, Saryazdi, Keetha, Tewari, Tenenbaum, de~Melo, Krishna, Paull, Shkurti, and Torralba]{jatavallabhula2023conceptfusion}
K.~M. Jatavallabhula, A.~Kuwajerwala, Q.~Gu, M.~Omama, T.~Chen, A.~Maalouf, S.~Li, G.~Iyer, S.~Saryazdi, N.~Keetha, A.~Tewari, J.~B. Tenenbaum, C.~M. de~Melo, M.~Krishna, L.~Paull, F.~Shkurti, and A.~Torralba.
\newblock Conceptfusion: Open-set multimodal 3d mapping.
\newblock In \emph{Robotics: Science and Systems (RSS)}, 2023.

\bibitem[Kerr et~al.(2023)Kerr, Kim, Goldberg, Kanazawa, and Tancik]{kerr2023lerf}
J.~Kerr, C.~M. Kim, K.~Goldberg, A.~Kanazawa, and M.~Tancik.
\newblock Lerf: Language embedded radiance fields.
\newblock In \emph{IEEE/CVF International Conference on Computer Vision (ICCV)}, pages 19672--19682, 2023.

\bibitem[Peng et~al.(2023)Peng, Genova, Jiang, Tagliasacchi, Pollefeys, and Funkhouser]{peng2022openscene}
S.~Peng, K.~Genova, C.~Jiang, A.~Tagliasacchi, M.~Pollefeys, and T.~Funkhouser.
\newblock Openscene: 3d scene understanding with open vocabularies.
\newblock In \emph{IEEE/CVF Conference on Computer Vision and Pattern Recognition (CVPR)}, pages 815--824, 2023.

\bibitem[Rana et~al.(2023)Rana, Haviland, Garg, Abou-Chakra, Reid, and Suenderhauf]{rana2023sayplan}
K.~Rana, J.~Haviland, S.~Garg, J.~Abou-Chakra, I.~Reid, and N.~Suenderhauf.
\newblock Sayplan: Grounding large language models using 3d scene graphs for scalable robot task planning.
\newblock In \emph{Conference on Robot Learning (CoRL)}, 2023.

\bibitem[Savva et~al.(2019)Savva, Kadian, Maksymets, Zhao, Wijmans, Jain, Straub, Liu, Koltun, Malik, Parikh, and Batra]{savva2019habitat}
M.~Savva, A.~Kadian, O.~Maksymets, Y.~Zhao, E.~Wijmans, B.~Jain, J.~Straub, J.~Liu, V.~Koltun, J.~Malik, D.~Parikh, and D.~Batra.
\newblock Habitat: A platform for embodied ai research.
\newblock \emph{Proceedings of the IEEE/CVF International Conference on Computer Vision (ICCV)}, 2019.

\bibitem[Parisotto and Salakhutdinov(2018)]{parisotto2018neuralmap}
E.~Parisotto and R.~Salakhutdinov.
\newblock Neural map: Structured memory for deep reinforcement learning.
\newblock In \emph{International Conference on Learning Representations (ICLR)}, 2018.

\bibitem[Parisotto et~al.(2020)Parisotto, Song, Rae, Pascanu, Gulcehre, Jayakumar, Jaderberg, Kaufman, Clark, Noury, Botvinick, Heess, and Hadsell]{parisotto2020stabilizing}
E.~Parisotto, F.~Song, J.~W. Rae, R.~Pascanu, C.~Gulcehre, S.~M. Jayakumar, M.~Jaderberg, R.~L. Kaufman, A.~Clark, S.~Noury, M.~Botvinick, N.~Heess, and R.~Hadsell.
\newblock Stabilizing transformers for reinforcement learning.
\newblock In \emph{International Conference on Machine Learning (ICML)}, 2020.

\bibitem[Baker et~al.(2022)Baker, Akkaya, Zhokhov, Huizinga, Tang, Ecoffet, Houghton, Sampedro, and Clune]{baker2022video}
B.~Baker, I.~Akkaya, P.~A. Zhokhov, J.~Huizinga, J.~Tang, A.~Ecoffet, B.~Houghton, R.~Sampedro, and J.~Clune.
\newblock Video pretraining (vpt): Learning to act by watching unlabeled online videos.
\newblock In \emph{Advances in Neural Information Processing Systems (NeurIPS)}, pages 24639--24654, 2022.

\bibitem[Liu et~al.(2024)Liu, Song, Jiang, Chen, Luo, Li, and Lin]{liu2024meia}
Y.~Liu, X.~Song, K.~Jiang, W.~Chen, J.~Luo, G.~Li, and L.~Lin.
\newblock Meia: Multimodal embodied perception and interaction in unknown environments.
\newblock \emph{arXiv preprint arXiv:2402.00290}, 2024.

\bibitem[Yang et~al.(2025)Yang, Yang, Zhou, Chen, Zhang, Du, and Gan]{yang20243dmem}
Y.~Yang, H.~Yang, J.~Zhou, P.~Chen, H.~Zhang, Y.~Du, and C.~Gan.
\newblock 3d-mem: 3d scene memory for embodied exploration and reasoning.
\newblock In \emph{IEEE/CVF Conference on Computer Vision and Pattern Recognition (CVPR)}, pages 17294--17303, 2025.

\bibitem[Xie et~al.(2024)Xie, Min, Ji, Yang, Zhang, Xu, Bajaj, Salakhutdinov, Johnson-Roberson, and Bisk]{xie2024embodiedrag}
Q.~Xie, S.~Y. Min, P.~Ji, Y.~Yang, T.~Zhang, K.~Xu, A.~Bajaj, R.~Salakhutdinov, M.~Johnson-Roberson, and Y.~Bisk.
\newblock Embodied-rag: General non-parametric embodied memory for retrieval and generation.
\newblock \emph{arXiv preprint arXiv:2409.18313}, 2024.

\bibitem[Kaelbling et~al.(1998)Kaelbling, Littman, and Cassandra]{kaelbling1998planning}
L.~P. Kaelbling, M.~L. Littman, and A.~R. Cassandra.
\newblock Planning and acting in partially observable stochastic domains.
\newblock \emph{Artificial Intelligence}, 101\penalty0 (1--2):\penalty0 99--134, 1998.
\newblock \doi{10.1016/S0004-3702(98)00023-X}.

\bibitem[Kochenderfer(2015)]{kochenderfer2015decision}
M.~J. Kochenderfer.
\newblock \emph{Decision Making Under Uncertainty: Theory and Application}.
\newblock MIT Press, Cambridge, MA, 2015.

\bibitem[Kochenderfer et~al.(2022)Kochenderfer, Wheeler, and Wray]{kochenderfer2022algorithms}
M.~J. Kochenderfer, T.~A. Wheeler, and K.~H. Wray.
\newblock \emph{Algorithms for Decision Making}.
\newblock MIT Press, Cambridge, MA, 2022.

\bibitem[Kaelbling and Lozano-Perez(2012)]{kaelbling2012unifying}
L.~P. Kaelbling and T.~Lozano-Perez.
\newblock Unifying perception, estimation and action for mobile manipulation via belief space planning.
\newblock In \emph{IEEE International Conference on Robotics and Automation (ICRA)}, pages 2952--2959, 2012.
\newblock \doi{10.1109/ICRA.2012.6225237}.

\bibitem[Kaelbling and Lozano-Perez(2013)]{kaelbling2013integrated}
L.~P. Kaelbling and T.~Lozano-Perez.
\newblock Integrated task and motion planning in belief space.
\newblock \emph{The International Journal of Robotics Research}, 32\penalty0 (9--10):\penalty0 1194--1227, 2013.
\newblock \doi{10.1177/0278364913484072}.

\bibitem[Garrett et~al.(2020)Garrett, Paxton, Lozano-Perez, Kaelbling, and Fox]{garrett2020online}
C.~R. Garrett, C.~Paxton, T.~Lozano-Perez, L.~P. Kaelbling, and D.~Fox.
\newblock Online replanning in belief space for partially observable task and motion problems.
\newblock In \emph{IEEE International Conference on Robotics and Automation (ICRA)}, pages 5678--5684, 2020.

\bibitem[Zhao et~al.(2025)Zhao, McClinton, Curtis, Kumar, Silver, Kaelbling, and Wong]{zhao2025seeing}
L.~Zhao, W.~McClinton, A.~Curtis, N.~Kumar, T.~Silver, L.~P. Kaelbling, and L.~L.~S. Wong.
\newblock Seeing is believing: Belief-space planning with foundation models as uncertainty estimators.
\newblock \emph{arXiv preprint arXiv:2504.03245}, 2025.

\bibitem[Mo et~al.(2021)Mo, Guibas, Mukadam, Gupta, and Tulsiani]{mo2021where2act}
K.~Mo, L.~Guibas, M.~Mukadam, A.~Gupta, and S.~Tulsiani.
\newblock Where2act: From pixels to actions for articulated 3d objects.
\newblock In \emph{IEEE/CVF International Conference on Computer Vision (ICCV)}, 2021.

\bibitem[Jiang et~al.(2022)Jiang, Hsu, and Zhu]{jiang2022ditto}
Z.~Jiang, C.-C. Hsu, and Y.~Zhu.
\newblock Ditto: Building digital twins of articulated objects from interaction.
\newblock In \emph{IEEE/CVF Conference on Computer Vision and Pattern Recognition (CVPR)}, 2022.

\bibitem[Ichter et~al.(2022)Ichter, Brohan, Chebotar, Finn, Hausman, Herzog, Ho, Ibarz, Irpan, Jang, Julian, Kalashnikov, Levine, Lu, Parada, Rao, Sermanet, Toshev, Vanhoucke, Xia, Xiao, Xu, Yan, Brown, Ahn, Cortes, Sievers, Tan, Xu, Reyes, Rettinghouse, Quiambao, Pastor, Luu, Lee, Kuang, Jesmonth, Joshi, Jeffrey, Ruano, Hsu, Gopalakrishnan, David, Zeng, and Fu]{ahn2022saycan}
B.~Ichter, A.~Brohan, Y.~Chebotar, C.~Finn, K.~Hausman, A.~Herzog, D.~Ho, J.~Ibarz, A.~Irpan, E.~Jang, R.~Julian, D.~Kalashnikov, S.~Levine, Y.~Lu, C.~Parada, K.~Rao, P.~Sermanet, A.~T. Toshev, V.~Vanhoucke, F.~Xia, T.~Xiao, P.~Xu, M.~Yan, N.~Brown, M.~Ahn, O.~Cortes, N.~Sievers, C.~Tan, S.~Xu, D.~Reyes, J.~Rettinghouse, J.~Quiambao, P.~Pastor, L.~Luu, K.-H. Lee, Y.~Kuang, S.~Jesmonth, N.~J. Joshi, K.~Jeffrey, R.~J. Ruano, J.~Hsu, K.~Gopalakrishnan, B.~David, A.~Zeng, and C.~K. Fu.
\newblock Do as i can, not as i say: Grounding language in robotic affordances.
\newblock In \emph{Conference on Robot Learning (CoRL)}, 2022.

\bibitem[Driess et~al.(2023)Driess, Xia, Sajjadi, Lynch, Chowdhery, Ichter, Wahid, Tompson, Vuong, Yu, Huang, Chebotar, Sermanet, Duckworth, Levine, Vanhoucke, Hausman, Toussaint, Greff, Zeng, Mordatch, and Florence]{driess2023palm}
D.~Driess, F.~Xia, M.~S.~M. Sajjadi, C.~Lynch, A.~Chowdhery, B.~Ichter, A.~Wahid, J.~Tompson, Q.~Vuong, T.~Yu, W.~Huang, Y.~Chebotar, P.~Sermanet, D.~Duckworth, S.~Levine, V.~Vanhoucke, K.~Hausman, M.~Toussaint, K.~Greff, A.~Zeng, I.~Mordatch, and P.~Florence.
\newblock Palm-e: An embodied multimodal language model.
\newblock In \emph{International Conference on Machine Learning (ICML)}, 2023.

\bibitem[Zitkovich et~al.(2023)Zitkovich, Yu, Xu, Xu, Xiao, Xia, Wu, Wohlhart, Welker, Wahid, Vuong, Vanhoucke, Tran, Soricut, Singh, Singh, Sermanet, Sanketi, Salazar, Ryoo, Reymann, Rao, Pertsch, Mordatch, Michalewski, Lu, Levine, Lee, Lee, Leal, Kuang, Kalashnikov, Julian, Joshi, Irpan, Ichter, Hsu, Herzog, Hausman, Gopalakrishnan, Fu, Florence, Finn, Dubey, Driess, Ding, Choromanski, Chen, Chebotar, Carbajal, Brown, Brohan, Arenas, and Han]{brohan2023rt2}
B.~Zitkovich, T.~Yu, S.~Xu, P.~Xu, T.~Xiao, F.~Xia, J.~Wu, P.~Wohlhart, S.~Welker, A.~Wahid, Q.~Vuong, V.~Vanhoucke, H.~Tran, R.~Soricut, A.~Singh, J.~Singh, P.~Sermanet, P.~R. Sanketi, G.~Salazar, M.~S. Ryoo, K.~Reymann, K.~Rao, K.~Pertsch, I.~Mordatch, H.~Michalewski, Y.~Lu, S.~Levine, L.~Lee, T.-W.~E. Lee, I.~Leal, Y.~Kuang, D.~Kalashnikov, R.~Julian, N.~J. Joshi, A.~Irpan, B.~Ichter, J.~Hsu, A.~Herzog, K.~Hausman, K.~Gopalakrishnan, C.~Fu, P.~Florence, C.~Finn, K.~A. Dubey, D.~Driess, T.~Ding, K.~M. Choromanski, X.~Chen, Y.~Chebotar, J.~Carbajal, N.~Brown, A.~Brohan, M.~G. Arenas, and K.~Han.
\newblock Rt-2: Vision-language-action models transfer web knowledge to robotic control.
\newblock In \emph{Conference on Robot Learning (CoRL)}, 2023.

\bibitem[Lin et~al.(2023)Lin, Agia, Migimatsu, Pavone, and Bohg]{lin2023text2motion}
K.~Lin, C.~Agia, T.~Migimatsu, M.~Pavone, and J.~Bohg.
\newblock Text2motion: From natural language instructions to feasible plans.
\newblock \emph{Autonomous Robots}, pages 1345--1365, 2023.

\bibitem[Huang et~al.(2023)Huang, Wang, Zhang, Li, Wu, and Fei-Fei]{huang2023voxposer}
W.~Huang, C.~Wang, R.~Zhang, Y.~Li, J.~Wu, and L.~Fei-Fei.
\newblock Voxposer: Composable 3d value maps for robotic manipulation with language models.
\newblock In \emph{Conference on Robot Learning (CoRL)}, 2023.

\bibitem[Hafner et~al.(2020)Hafner, Lillicrap, Ba, and Norouzi]{hafner2020dreamer}
D.~Hafner, T.~Lillicrap, J.~Ba, and M.~Norouzi.
\newblock Dream to control: Learning behaviors by latent imagination.
\newblock \emph{International Conference on Learning Representations (ICLR)}, 2020.

\bibitem[Carion et~al.(2025)Carion, Gustafson, Hu, Debnath, Hu, Suris, Ryali, Alwala, Khedr, Huang, Lei, Ma, Guo, Kalla, Marks, Greer, Wang, Sun, Rädle, Afouras, Mavroudi, Xu, Wu, Zhou, Momeni, Hazra, Ding, Vaze, Porcher, Li, Li, Kamath, Cheng, Dollár, Ravi, Saenko, Zhang, and Feichtenhofer]{carion2025sam3segmentconcepts}
N.~Carion, L.~Gustafson, Y.-T. Hu, S.~Debnath, R.~Hu, D.~Suris, C.~Ryali, K.~V. Alwala, H.~Khedr, A.~Huang, J.~Lei, T.~Ma, B.~Guo, A.~Kalla, M.~Marks, J.~Greer, M.~Wang, P.~Sun, R.~Rädle, T.~Afouras, E.~Mavroudi, K.~Xu, T.-H. Wu, Y.~Zhou, L.~Momeni, R.~Hazra, S.~Ding, S.~Vaze, F.~Porcher, F.~Li, S.~Li, A.~Kamath, H.~K. Cheng, P.~Dollár, N.~Ravi, K.~Saenko, P.~Zhang, and C.~Feichtenhofer.
\newblock Sam 3: Segment anything with concepts, 2025.
\newblock URL \url{https://arxiv.org/abs/2511.16719}.

\bibitem[Todorov et~al.(2012)Todorov, Erez, and Tassa]{todorov2012mujoco}
E.~Todorov, T.~Erez, and Y.~Tassa.
\newblock Mujoco: A physics engine for model-based control.
\newblock In \emph{2012 IEEE/RSJ International Conference on Intelligent Robots and Systems}, pages 5026--5033. IEEE, 2012.
\newblock \doi{10.1109/IROS.2012.6386109}.

\bibitem[Nasiriany et~al.(2026)Nasiriany, Nasiriany, Maddukuri, and Zhu]{robocasa365}
S.~Nasiriany, S.~Nasiriany, A.~Maddukuri, and Y.~Zhu.
\newblock Robocasa365: A large-scale simulation framework for training and benchmarking generalist robots.
\newblock In \emph{International Conference on Learning Representations (ICLR)}, 2026.

\bibitem[Wu et~al.(2024)Wu, Chong, Holmberg, Prasad, Gao, Khatib, Song, Rusinkiewicz, and Bohg]{wu2024tidybotpp}
J.~Wu, W.~Chong, R.~Holmberg, A.~Prasad, Y.~Gao, O.~Khatib, S.~Song, S.~Rusinkiewicz, and J.~Bohg.
\newblock Tidybot++: An open-source holonomic mobile manipulator for robot learning, 2024.
\newblock URL \url{https://arxiv.org/abs/2412.10447}.

\bibitem[Kapfer et~al.(2025)Kapfer, Stine, Narasimhan, Mentzel, and Candès]{marlowe2025}
C.~Kapfer, K.~Stine, B.~Narasimhan, C.~Mentzel, and E.~Candès.
\newblock Marlowe: Stanford's gpu-based computational instrument, 2025.
\newblock URL \url{https://doi.org/10.5281/zenodo.14751899}.

\bibitem[Schaal(2006)]{schaal2006dmp}
S.~Schaal.
\newblock Dynamic movement primitives: A framework for motor control in humans and humanoid robotics.
\newblock In H.~Kimura, K.~Tsuchiya, A.~Ishiguro, and H.~Witte, editors, \emph{Adaptive Motion of Animals and Machines}, pages 261--280. Springer, Tokyo, Japan, 2006.
\newblock \doi{10.1007/4-431-31381-8_23}.

\bibitem[Sundaresan et~al.(2025)Sundaresan, Malhotra, Miao, Yang, Wu, Hu, Antonova, Engelmann, Sadigh, and Bohg]{sundaresan2025homerlearninginthewildmobile}
P.~Sundaresan, R.~Malhotra, P.~Miao, J.~Yang, J.~Wu, H.~Hu, R.~Antonova, F.~Engelmann, D.~Sadigh, and J.~Bohg.
\newblock Homer: Learning in-the-wild mobile manipulation via hybrid imitation and whole-body control, 2025.
\newblock URL \url{https://arxiv.org/abs/2506.01185}.

\bibitem[Yang et~al.(2023)Yang, Zhang, Li, Zou, Li, and Gao]{yang2023setofmarkpromptingunleashesextraordinary}
J.~Yang, H.~Zhang, F.~Li, X.~Zou, C.~Li, and J.~Gao.
\newblock Set-of-mark prompting unleashes extraordinary visual grounding in gpt-4v, 2023.
\newblock URL \url{https://arxiv.org/abs/2310.11441}.

\end{thebibliography}

\newpage
\appendix
\appendix
\section*{Appendix}
This appendix provides additional implementation details, full prompts, memory structure, and scenario descriptions for \ApproachName{}.

\section{Implementation Details}
\label{app:implementation}

\subsection{Experimental Setup}
\label{app:exp-setup}

Each trial consists of a sequence of natural-language instructions revealed one at a time. This requires the robot to reuse information gathered from earlier observations and interactions rather than solving each instruction from scratch. At each planner call, the robot receives the current scene graph, any retrieved keyframes, and the available primitive skills. It selects a high-level action, executes the first step, updates memory when applicable, and then replans from the updated state.

In simulation, tasks are evaluated with a fixed budget of twelve planner calls. A task is successful if the requested instruction is completed within this budget. For sequential tasks, we report the fraction of subtasks completed and the number of high-level steps used.

\paragraph{Simulation.}
We evaluate in RoboCasa365~\citep{robocasa365} kitchen environments with custom memory-intensive task families. The simulated robot is a mobile manipulator with a Franka Panda arm and an Omron mobile base. RGB-D observations are rendered from multiple third-person cameras for scene graph construction, while the wrist camera is used for interaction analysis and keyframes. In scenarios that require prior exploration, the robot first observes or interacts with relevant fixtures to seed memory before receiving the test instruction.

Task success is evaluated from environment state and execution logs rather than from the planner's self-report. Depending on the scenario, success may require opening the correct fixture, discovering the requested object, placing an object in the correct location, terminating when the goal is already satisfied, or correctly answering a recall-style instruction from memory.

\paragraph{Real world.}
Our real-world implementation runs on a mobile manipulation platform with a 6-DoF arm, parallel-jaw gripper, RGB-D perception cameras, and a wrist camera. The system maintains an object-centric scene graph online during execution. Perception updates add or refresh objects in the graph, while interaction updates add discovered contents, object properties, task outcomes, and linked keyframes. The real-world tasks are operator-gated for safety, and final task success is operator-confirmed. At the start of each trial the robot's base orientation is initialized to face forward toward the workspace.

\subsection{High-Level Primitive Skills}
\label{app:primitives}

All methods share the same high-level primitive interface. The low-level controllers differ between simulation and the real robot, but each primitive exposes the same kind of outcome information to the planner and, when enabled, to the interaction analyzer.

\paragraph{Simulation primitives.}
In simulation, navigation and inspection are implemented with scripted helpers, while door manipulation uses closed-loop policies when available.

\begin{itemize}
    \item \texttt{move\_base\_to(fixture\,/\,pos)}: move the base to a fixture or explicit target pose.
    \item \texttt{move\_arm\_to(pos)}: move the end effector to a target Cartesian position.
    \item \texttt{OpenCabinet(fixture)}: Opens a target fixture. The primitive automatically opens the fixture closest to the robot's current location. To capture the dynamic interaction in the Locked/Unlocked Cabinet case, we train a learned policy to open the cabinet. 
    \item \texttt{PickPlaceCabinetToCounter(object)}: pick an object from the cabinet and place it onto a counter. Ground-truth object location is used for the primitive.
    \item \texttt{Inspect(fixture, object)}: move the wrist camera through an open fixture to observe its contents.
\end{itemize}

After each non-motion primitive, the interaction analyzer can inspect the action window and write a success/failure outcome into memory. Motion primitives instead report whether they reached the requested tolerance.

\paragraph{Real-world primitives.}
On the real robot, manipulation skills are implemented as dynamic movement primitives (DMPs)~\citep{schaal2006dmp} fit from a small number of teleoperated demonstrations. These include navigation to a selected location, opening and closing drawers, peeking into a fixture, picking up an object, placing an object, and moving to a remembered location. For object-directed manipulation, we use a 3D pointing module following the procedure in \citet{sundaresan2025homerlearninginthewildmobile}. Given the current camera image, a VLM points to a task-relevant grasp location, such as a drawer handle or the body of the target object. This 2D point is back-projected through the aligned depth map into a 3D pre-grasp target in the robot frame, which shifts the demonstrated DMP rollout to the current object location. This lets a single demonstration generalize across nearby object placements. Primitive execution is operator-gated for safety. Interaction-based outcomes such as discovered contents or inferred fixture locked state are updated by the interaction analyzer.

\subsection{Scene Graph Construction}
\label{app:scene-graph}

\paragraph{SAM3 prompts.}
\label{app:sam3-prompts}

We run SAM3 \cite{carion2025sam3segmentconcepts} with a scenario-specific prompt set shared across methods, so differences in performance come from the memory system rather than perception vocabulary. The object vocabulary is derived from the task setting, while fixture prompts cover the containers and surfaces relevant for mobile manipulation. For real-world experiments, several object categories (e.g., gaming controller, scissors) present in the experiments are intentionally left out to illustrate the limitation of a closed-set scene graph.

\begin{table}[h]
\centering
\small
\setlength{\tabcolsep}{4pt}
\begin{tabular}{ll}
\toprule
\textbf{Group} & \textbf{SAM3 prompts} \\
\midrule
Objects (sim) & categories present in the scenario (e.g., ketchup, \\
& mayonnaise, mustard, cereal, spaghetti box, lemon, pear, \dots) \\
Fixtures (sim, perception-fixture mode) & cabinet, drawer, counter, oven, microwave, \\
& sink, refrigerator, dishwasher, stove \\
Real-world objects & drawer, cup, chair, sock\\
\bottomrule
\end{tabular}
\caption{SAM3 prompts used for scene graph construction.}
\label{tab:sam3-prompts}
\end{table}

\paragraph{Node association and update.}
\label{app:sg-params}

Detections with the same semantic label and nearby 3D positions are merged into object-centric scene graph nodes. New observations are associated to existing same-label nodes using spatial proximity, and matched nodes update their centroid with a running average. Detections that do not match any existing node initialize new graph nodes. Object identities reset between trials.

\paragraph{Node format.}
\label{app:sg-node-format}

Scene graph nodes are initialized with spatial and semantic fields. Interaction-derived properties, contents, and task history are attached as the robot acts in the environment. Keyframes are linked to nodes through a separate keyframe index.

\begin{GrayVerbatim}
{
 "cabinet_3": {
   "label": "cabinet",
   "pos": [x, y, z],
   "first_seen_frame": 124,
   "bbox_3d": [...],
   "properties": {
     "accessible": {"value": "unknown", "confidence": 0.5, "source": "perception"},
     "pickable":   {"value": "unknown", "confidence": 0.5, "source": "perception"},
     "pushable":   {"value": "unknown", "confidence": 0.5, "source": "perception"},
     "fullness":   {"value": "unknown", "confidence": 0.3, "source": "perception"},
     "openable":   {"value": true, "confidence": 0.9, "source": "interaction",
                    "is_opened": true, "reasoning": "Door visibly rotates open."}
   },
   "contents": [{"name": "banana", "value": "...", "reasoning": "..."}],
   "contents_empty": false,
   "contents_source_action": "Inspect",
   "task_history": [{"task": "OpenCabinet", "success": true}]
 }
}
\end{GrayVerbatim}

\section{Prompting and Memory Update Details}
\label{app:prompts-memory}

\subsection{VLM Interaction Analyzer}
\label{app:interaction-analyzer}

After each manipulation primitive, the interaction analyzer receives a compact visual summary of the action window, together with the action name, target node, current task, and the node's prior memory state. The analyzer updates memory only when the visual evidence supports a change, such as a fixture opening, an object being lifted, or contents becoming visible.

\paragraph{Collage construction.}
For each interaction, we provide the analyzer with frames from before, during, and after the action. The before and after images preserve high-resolution endpoint evidence, while the action collage summarizes the gripper engagement window. This allows the analyzer to distinguish transient execution issues, such as a missed grasp, from persistent object properties, such as a locked drawer.

\begin{figure}[h]
\centering
\includegraphics[width=0.95\linewidth]{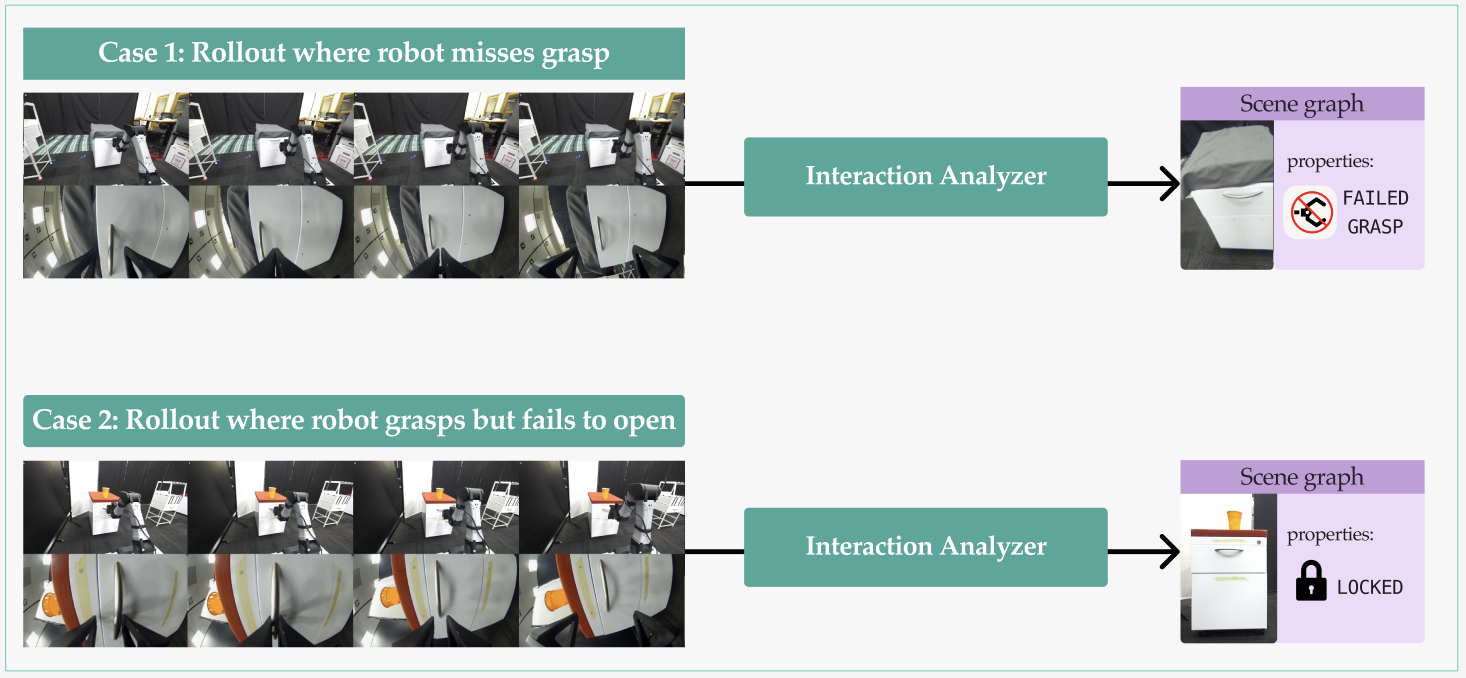}
\caption{
\textbf{VLM Interaction Analyzer.} The interaction analyzer reasons over a collage of frames collected during the interaction, capturing dynamic cues that are only visible as the action is taken. \ApproachName{}'s interaction analyzer is able to differentiate between a failed grasp in Case 1, where the robot misses the handle, and a locked cabinet in Case 2, where the robot grasps and pulls but the drawer does not open.
}
\label{fig:interaction-analyzer-fig}
\end{figure}

\paragraph{Interaction analyzer prompt.}
\label{app:ia-prompt}

The full interaction analyzer prompt is shown below. The prompt instructs the VLM to reason over the full action window, separate execution failure from object properties, update only with visual support, and return a strict JSON object.

\begin{GrayVerbatim}
You are an object belief updater for a mobile manipulation robot.

Current belief state of the target object:
{json belief state of the target scene-graph node}

The robot attempted the action: "{action_name}" targeting position {target_pos}.

Terminology: a "fixture" is any scene element with an interior that can be
opened to reveal contents.

You will receive THREE images: (1) full-resolution BEFORE (arm at rest, not a
grasp-attempt frame); (2) full-resolution AFTER (arm auto-reset to rest - the
withdrawal is by design and is NOT evidence of a failed grasp; compare with
BEFORE for fixture angle change); (3) a multi-camera collage of the MANIPULATION
WINDOW only (pre-position approach and auto-reset excluded). Rows show different
camera views of the same window at the same timesteps {rows}; columns are time,
left-to-right, labeled BEFORE..AFTER. Use column-wise agreement across rows to
disambiguate (e.g. the wrist close-up confirms whether the gripper engaged the
handle visible in the base view).

Use the action name to pick the analysis: MANIPULATION (BEFORE/AFTER bracket the
interaction; compare to detect what moved/opened/closed/displaced, and report any
revealed interior) vs INSPECTION (BEFORE/AFTER are different wrist poses of the
same interior; aggregate to identify contents; apparent motion is parallax).
Interiors can be dark - check the full-resolution images.

[Nearby fixture nodes in the scene graph, with exact IDs to use in
contents_updates and any prior contents listed as a hypothesis to merge.]

Properties of interest (when the action moves/manipulates the target):
- accessible (true/false): can the robot reach it?
- pickable (true/false): only true if FULLY LIFTED off surface.
- pushable (true/false): can it be pushed/slid to a new position?
- fullness (empty / partially_full / full / unknown)
- openable (true/false), distinguishing three cases:
   (a) opened (any angular change) -> openable=true, is_opened=true
   (b) no motion despite a genuine attempt -> describe as locked/stuck so the
       planner moves to a different fixture
   (c) the attempt failed (no firm contact) -> describe as failed grasp so the
       planner can retry
 Procedure to separate (b) from (c): scan the WRIST row across ALL columns - a
 close-up of the handle/drawer face means contact -> (b). Else scan SIDE/FRONT
 rows for jaws CLOSED ON the handle -> (b). Only if BOTH find zero contact
 evidence -> (c). If ambiguous, PREFER (b). A locked drawer can shear the jaws
 sideways off the handle, so side-view offset is not evidence of (c) - check
 the wrist row at the same column first.

Analysis: (1) PHYSICAL CHANGE - track the target across collage frames: shifting
or stationary despite contact? Update accessible/pickable/pushable/fullness as
warranted (inspections: expect no real change). (2) VISIBLE CONTENTS - identify
objects visibly INSIDE any revealed fixture interior, labeled as specifically as
the evidence allows; one contents_updates entry per fixture using the exact SG
id; empty list if no interior is visible.

Task success (by intent): push/slide -> target visibly displaced (pushable=true);
pick/lift -> target FULLY LIFTED; open/close -> fixture in requested state in
AFTER (set openable, is_opened); inspection -> a usable view was obtained.

Rules: only update with clear visual evidence; attribute contents only to the
fixture they are visibly inside; [] when no interior is visible; one short
reasoning per update; do not hallucinate properties or contents.

Return STRICT JSON ONLY:  { ... output schema shown below ... }
\end{GrayVerbatim}

\paragraph{Property updater and outcome schema.}
\label{app:updater}

The analyzer writes structured memory updates back to the scene graph. These include task success, failure reason, property updates, newly discovered open-vocabulary properties, and fixture contents. The seeded property vocabulary includes \texttt{accessible}, \texttt{pickable}, \texttt{pushable}, \texttt{fullness}, and \texttt{openable}. The \texttt{is\_opened} field is used only for \texttt{openable} updates.

\begin{GrayVerbatim}
{
 "task_success": true,
 "failure_reason": null,
 "property_updates": [
   {"object_id": "cabinet_3", "property": "openable", "old_value": "unknown",
    "new_value": true, "confidence": 0.9, "reasoning": "Door rotates open.",
    "is_opened": true}
 ],
 "new_properties": [
   {"name": "has_handle", "value": true, "reasoning": "Pull handle visible."}
 ],
 "contents_updates": [
   {"fixture_id": "cabinet_3",
    "contents": [{"name": "banana", "description": "..."},
                 {"name": "mug", "description": "..."}],
    "is_empty": false, "reasoning": "Opened cabinet reveals a banana and mug."}
 ],
 "confidence": 0.8
}
\end{GrayVerbatim}

\paragraph{Failed grasp vs.\ locked fixture.}
\label{app:failed-vs-locked}

A key role of the analyzer is to separate execution failures from persistent object properties. As shown in Fig.~\ref{fig:interaction-analyzer-fig}, a missed or slipped grasp is recorded as an execution failure, allowing the planner to retry. In contrast, a firm pull with no motion is recorded as evidence that the fixture is locked or stuck, allowing the planner to avoid repeating the same failed interaction.

\subsection{Keyframe Memory}
\label{app:keyframes}

Keyframes preserve visual details that are difficult to compress into scene graph fields, such as clutter, object arrangement, and visual distinctions between similar objects. They are linked to scene graph nodes through a keyframe index, allowing later tasks to retrieve visual evidence associated with relevant objects, fixtures, or interactions.

\paragraph{Sources.}
\label{app:kf-sources}

Keyframes are saved from both interaction events and perception events. Interaction keyframes include frames before, during, and after a manipulation, as well as analyzer-selected peak frames when an interaction reveals useful contents while perception-side keyframes capture new significant scene graph changes.

\begin{table}[h]
\centering
\small
\setlength{\tabcolsep}{4pt}
\begin{tabular}{lll}
\toprule
\textbf{Trigger} & \textbf{Source} & \textbf{Purpose} \\
\midrule
\texttt{interaction\_pre} & Manipulation & State before action \\
\texttt{interaction\_during} & Manipulation & Mid-action evidence \\
\texttt{interaction\_post} & Manipulation & State after action \\
\texttt{interaction\_peak} & Analyzer & Informative reveal frame \\
\texttt{scene\_graph\_delta} & Perception & Significant graph update \\
\bottomrule
\end{tabular}
\caption{Sources of saved keyframes.}
\label{tab:kf-sources}
\end{table}

\paragraph{Metadata.}
\label{app:kf-metadata}

Each keyframe stores metadata that makes it searchable and linkable to the scene graph: timestamp, camera, trigger source, parent interaction when applicable, linked scene graph nodes, robot pose, labels, caption, and image paths.

\begin{GrayVerbatim}
{
 "frame_id": "a1b2c3d4e5f6...",
 "timestamp": 172.4,
 "session_id": "20260603_142210",
 "camera": "base1",
 "trigger": "interaction_peak",
 "action": "OpenCabinet",
 "parent_frame": "<pre-frame id or null>",
 "scene_graph_nodes": ["cabinet_3", "banana_1"],
 "robot_pose": [x, y, theta],
 "tags": ["banana", "mug"],
 "labels": [{"label": "cabinet interior", "source": "analyzer"},
            {"label": "banana", "source": "analyzer"}],
 "caption": "...",
 "image_path": "...",
 "thumb_path": "..."
}
\end{GrayVerbatim}

\paragraph{Duplicate filtering.}
To keep the keyframe memory compact, the system filters out blank, blurry, and near-duplicate frames before saving perception-triggered keyframes. Interaction-triggered keyframes are preserved even if visually similar, since pre-, during-, post-, and peak-interaction frames may encode outcome-relevant evidence.

\subsection{VLM Retriever and Planner}
\label{app:retriever-planner-prompts}

The retriever selects task-relevant keyframes given the current task, scene graph, and keyframe metadata. It first filters candidate frames using textual metadata and scene graph, then reranks candidate thumbnails with a VLM. The planner receives the retrieved frames alongside the current scene graph, current image, execution history, and list of motion primitives.

\begin{figure}[h]
\centering
\includegraphics[width=0.95\linewidth]{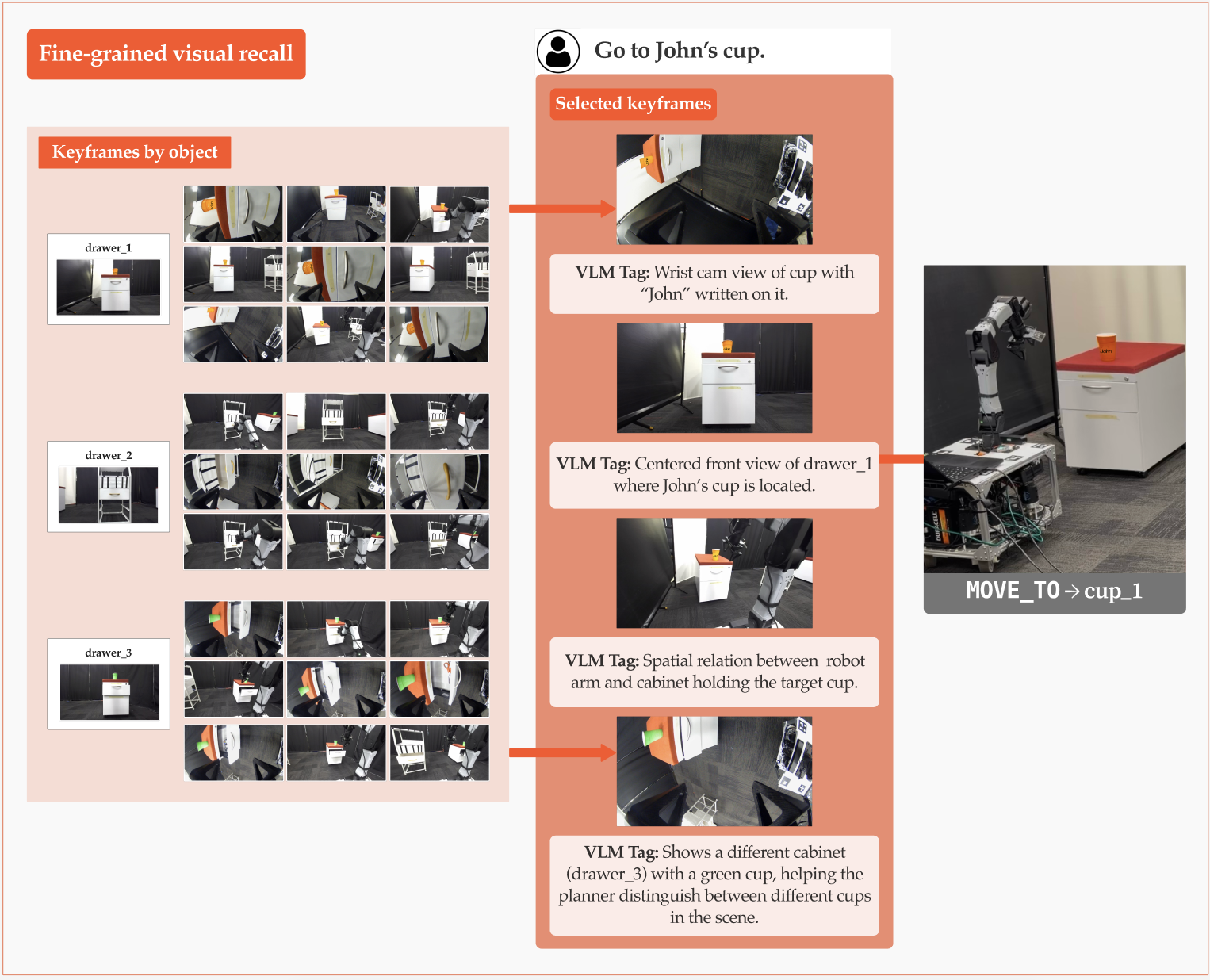}
\caption{
\textbf{VLM retriever effectively selects a task-relevant subset of keyframes for planning.} Given the scene graph and task description, the VLM retriever selects keyframes that are useful for the current goal and tags each one with a short reason. In this example, the task is to go to John's cup: the retrieved frames show both the orange cup on the left drawer and the green cup on the right drawer, allowing the planner to identify the orange cup labeled ``John'' as the correct target.
}
\label{fig:keyframes-retrieval}
\end{figure}

\paragraph{Retriever prompt.}
\label{app:retriever-prompt}

The full retrieval prompt is shown below. The no-scene-graph ablation removes the scene graph block and graph-node column.

\begin{GrayVerbatim}
You are a keyframe retrieval agent for a robot planner.

The robot has been given this task: "{goal}"
Here is the current scene graph (objects the planner knows about):
{scene_graph_summary}

Below is a catalog of stored keyframes. Each row: trigger, camera, action, the
scene-graph nodes it depicts, and any analyzer/instruction labels.
{catalog}    // idx | frame_id | trigger | camera | action | graph_nodes | labels

Choose up to {max_k} keyframes that would most help the planner. Consider: frames
showing objects/locations in the task; container interiors the planner must
reason about; spatial relationships (behind, inside, on top of). Return [] if none
help.

Return STRICT JSON ONLY:
{"chosen": [{"idx": <i>, "frame_id": "<first 12 chars>",
            "reason": "<one line on why this frame helps>"}]}
\end{GrayVerbatim}

\paragraph{Planner prompt.}
\label{app:planner-prompt}

We show the prompt for our full method (the \texttt{scene\_graph+kf} configuration; the baseline ablations strip the corresponding scene-graph fields and rules). The planner is called repeatedly in closed loop: it returns a multi-step plan, the system executes only the first step, then re-queries with updated observations and execution history. The current camera image is provided alongside the text, and retrieved keyframes are appended as additional images captioned with their trigger, camera, and retrieval reason. 

\medskip
\noindent\textbf{Simulation planner (RoboCasa kitchen).}
\begin{GrayVerbatim}
You are a task planner for a PandaOmron mobile manipulation robot in a RoboCasa
kitchen. The robot has a Panda arm mounted on an Omron mobile base with an
eye-in-hand wrist camera.

Goal: "{goal}"

{execution_history}     // prior attempts in order, each with SUCCESS/FAILED status

Available primitives (you may ONLY use these):
{primitives}            // name: description / Args / When

Kitchen fixtures in the scene (interactable objects the robot can navigate to):
{fixtures}

Scene graph (objects with 3D world positions, interaction-informed properties, and
inspection results including any "contents" found inside containers):
{scene_graph}

You are also given an image showing the current view of the scene. Use both the
scene graph data and the visual information to reason about spatial layout,
obstacles, and plan feasibility.

Robot physical constraints:
- The Panda arm has ~0.85m reach from the base centre.
- The Omron base is ~0.55m wide - it cannot fit through gaps narrower than 0.6m.
- Use move_base_to with a fixture name (target_fixture) to navigate the robot in
 front of any fixture; the system computes the approach position and heading.
- Use move_base_to with explicit target_pos [x, y, yaw] for non-fixture locations.
- Use move_arm_to for fine EEF placement before manipulation tasks.
- Always move_base_to the target fixture BEFORE attempting OpenCabinet, PickPlace,
 or Inspect - the arm must be in range of the target.
- Inspect performs a wrist-camera sweep across an open fixture, writing object
 identifications into scene_graph[fixture]["contents"]. Decide goal satisfaction
 from these contents (and any retrieved memory frames showing the fixture).

Generate a step-by-step plan to achieve the goal. You will be called repeatedly in a
closed loop: you produce a plan, the system executes the FIRST step, then calls you
again with updated observations and execution history. So plan ahead, but know that
you will get a chance to adapt after every step.

Rules:
1. ONLY use actions from the available primitives list.
2. For fixture-targeted primitives, use the exact fixture name from the fixtures list.
3. Order steps logically based on what the goal requires.
4. For objects that have properties (e.g. pickable, pushable, accessible, fullness,
  weight), leverage them to decide WHETHER to interact and which action to use.
5. Objects may have a task_history showing past attempts and their success/failure -
  use this to inform action choices. If an action failed on an object, do NOT retry
  a similar action on the same object - adapt your strategy.
6. Reference objects by their scene graph ID (e.g. banana_10, ketchup_3).
7. For move_base_to, provide target_fixture OR target_pos as [x, y, yaw_radians].
  For move_arm_to, provide target_pos as [x, y, z]. For all other primitives,
  provide target_fixture as the fixture name string.
8. For Inspect steps, set target_object to the object label the goal asks to find
  (e.g. "banana"). Use null for other actions.
9. After opening a container (cabinet, fridge, dishwasher) with an Open* primitive,
  always plan a corresponding Close* step before moving on or finishing the plan.
10. The goal is achieved once its literal request is satisfied. Informational verbs
   (find / locate / identify / where-is / describe / open) are satisfied as soon as
   the relevant state is observable from task_history, scene graph, or current image
   - no further physical interaction is required. Navigation verbs are satisfied as
   soon as task_history contains ONE successful move_base_to on the target fixture.
   Action verbs (pick up / place / bring) require the literal physical action. If the
   goal asks a question, your `assumptions` must contain the literal answer before
   returning an empty plan. If the goal is already achieved, return an EMPTY plan:
   {"plan": [], "assumptions": ["Goal already achieved."], "warnings": []}.

Return STRICT JSON ONLY:
{
 "plan": [
   {
     "step": 1,
     "action": "<primitive_name>",
     "target_fixture": "<fixture_name or null>",
     "target_pos": [x, y, z_or_yaw],
     "target_object": "<object label to find, or null>",
     "reasoning": "..."
   }
 ],
 "assumptions": ["..."],
 "warnings": ["..."]
}
\end{GrayVerbatim}

\noindent\textbf{Real-world planner.}
\begin{GrayVerbatim}
You are a task planner for a mobile manipulation robot.

Goal: "{goal}"

{execution_history}     // prior attempts in order, each with SUCCESS/FAILED status

Robot physical constraints:
- The robot base footprint is approximately 0.55m wide x 0.51m deep.
- The move_to action drives in a STRAIGHT LINE toward a target [x, y, z] position.
- The arm workspace is limited to +/-0.28m in the Y (left-right) direction from the
 base center.
- The arm can reach objects roughly 0.3-0.5m in front of the base (X direction).
- To interact with an object (open, close, peek), the robot must be within arm reach
 of it.

Scene graph (objects with 3D world positions, interaction-informed properties, and
inspection results including any "contents" found inside containers):
{scene_graph}

You are also given an image showing the current view of the scene. Use both the
scene graph data and the visual information to reason about spatial layout,
obstacles, and plan feasibility.

Image overlays:
- Some images include yellow circles with scene-graph node IDs (e.g. `drawer_1`)
 drawn at the projected positions of current SG nodes, geometrically projected
 from the current scene graph using known camera calibration.
- When an overlay appears in an image, prefer it as the authoritative ID mapping for
 that pixel region. If a memory frame's text note mentions a different ID than the
 overlay shows, defer to the overlay.
- Overlays appear only on the calibrated base frames. Wrist-camera frames pass
 through unannotated - for those, use the frame's text note as your grounding signal.
- An image may have no overlays even when one was expected (robot pose at capture was
 unknown). In that case, ground using visual + scene-graph reasoning as usual.

Available primitive actions:
{primitives}            // name: description, per registered primitive

Generate a step-by-step plan to achieve the goal. You will be called repeatedly in a
closed loop: you produce a plan, the system executes the FIRST step, then calls you
again with updated observations and execution history. So plan ahead, but know that
you will get a chance to adapt after every step.

Rules:
1. For objects that are in the path, leverage their properties (e.g. pickable,
  pushable, accessible, fullness, weight) to decide WHETHER to interact and which
  action to use.
2. Before issuing an action, check the target object's current properties in the
  scene graph to avoid redundant commands. For example, if is_opened is already
  true, consider whether you really need to re-issue an open command. You may still
  re-issue if the prior attempt only partially succeeded, but prefer skipping if the
  state already matches the goal.
3. Objects may have a task_history showing past attempts and their success/failure -
  use this to inform action choices, including whether to retry, adapt the approach,
  or try a different target.
  A drawer is only interactable if it has a visible handle the robot can grasp.
  Before planning ANY actions on a drawer (including move_to), verify from the images
  that it has a handle. A drawer with no handle is not openable - treat it as if it
  does not exist.
  If an open_drawer or close_drawer step fails, distinguish two cases: (a) the robot
  grasped the handle correctly but the drawer did not move (e.g. it is locked) -
  treat the entire drawer unit as non-openable and move on to a completely different
  drawer or fixture. Do NOT attempt to open a different drawer in the same unit;
  (b) the robot missed the handle or approached from a bad angle - retry the same
  action once, since visual pointing can produce a bad initial pose on the first
  attempt.
  When a new goal implies the environment has changed since prior inspections (e.g.
  "a new item was added"), the `contents` recorded in the scene graph may be STALE.
  Do not assume prior contents are still accurate. Use the goal description, scene
  graph properties, and task_history to decide which objects are worth re-inspecting
  - just the ones that are plausible given what you know and the task description.
4. When deciding where to put an object, refer to the scene graph contents and
  keyframe images of container interiors. Consider what items are already inside each
  container and where the object best belongs.
5. The robot can ONLY interact (open, close, peek) with the object it last did
  move_to on. Find the LAST move_to in the most recent goal in the execution history
  - if it differs from your target, plan a move_to first.
6. Reference objects by their scene graph ID (e.g. green_cube_0, red_cube_1).
7. ONLY use actions from the available primitives list.
8. Each step must have a target_object that exists in the scene graph (use the exact
  ID).
9. After inspecting or exploring, consider actions that increase executability of
  future steps (e.g. close an upper drawer to access a lower one).
10. Goal completion depends on the verb type. Informational verbs (find / locate /
   identify / where-is / describe / open) are satisfied as soon as the relevant state
   is observable from task_history, the scene graph, or current image - no further
   physical interaction with the target is required. Navigation verbs (navigate /
   go to / drive to / approach) are satisfied as soon as task_history contains ONE
   move step on the target or on the scene graph node nearest to / containing the
   target - do NOT re-issue the same navigation. Action verbs (pick up / place /
   bring) require the literal physical action. If the goal asks a question (describe
   / name / list / how many / which), your `reasoning` must contain the literal
   answer before you return an empty plan. If the goal is already achieved, return an
   EMPTY plan: {"plan": [], "reasoning": ["Goal already achieved."], "warnings": []}.
11. Some primitives are localized by VLM pointing on a camera frame. For those steps,
   you may include an optional `pointing_description` field - a short visual phrase
   describing exactly WHERE on the target the action should engage. Use it when the
   scene_graph_id is ambiguous (e.g. one `drawer_0` node represents a stack of
   multiple visible drawers). Examples: "the upper drawer handle", "the left cabinet
   door handle". Omit when the SG label alone is unambiguous.

Return STRICT JSON ONLY:
{
 "plan": [
   {
     "step": 1,
     "action": "<one of the available primitives>",
     "target_object": "<scene_graph_id>",
     "pointing_description": "<optional visual phrase; omit if unneeded>",
     "grounded_by_keyframe_idx": <optional int; 1-indexed memory frame used to
                                  identify target_object; omit when identified from
                                  the current scene only>,
     "reasoning": "..."
   }
 ],
 "reasoning": ["..."],
 "warnings": ["..."]
}
\end{GrayVerbatim}

\paragraph{Projecting scene graph nodes into images.}
\label{app:projection}

In real-world experiments, we use Set-of-Mark annotations~\citep{yang2023setofmarkpromptingunleashesextraordinary} to spatially ground the planner on calibrated base-camera images. For each visible scene graph node, we project its 3D position into the image and overlay a circle with the object label. Wrist-camera keyframes are left unannotated and are grounded through their captions and linked scene graph metadata.

\section{Simulation Scenario Details}
\label{app:sim-benchmark}

\subsection{Scenario Descriptions}
\label{app:sim-scenarios}

\begin{figure}[H]
\centering
\includegraphics[width=0.95\linewidth]{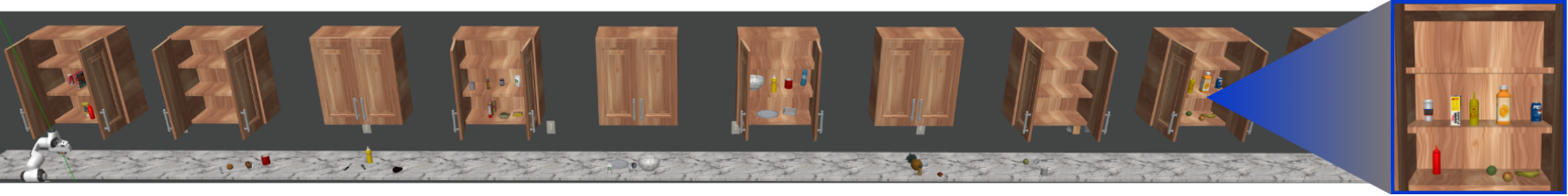}
\caption{
\textbf{25-task Long Horizon.} The 25-task sequence takes place in a ten-cabinet kitchen with
three locked cabinets and \textbf{more than 60 objects and fixtures to track.}
Countertop objects provide distractors and visual references.
The inset shows the contents and spatial arrangement of one cabinet.
}
\label{fig:sim-scenarios-25-task-long-horizon}
\end{figure}

\textbf{25-task Long Horizon.}
We evaluate an ordered sequence of 25 household tasks in a 10-cabinet
RoboCasa365 kitchen containing more than 60 objects and fixtures. Three
cabinets are locked, and objects on the counters provide distractors and
visual references. An initial search for a teapot in the final cabinet
encourages exploration; subsequent tasks reuse earlier observations and
interaction outcomes without resetting memory or the environment. The
sequence spans four task families:

\begin{itemize}
\item \textbf{Finding (9 tasks).}
Locate objects using remembered locations, neighboring items, or
accessibility. Examples include distinguishing a stored mug from one on
a counter, identifying ketchup by its surrounding condiments or fruit,
and finding the bottle with clear space for grasping.

\item \textbf{Matching (9 tasks).}
Distinguish three visual variants within each of seven object categories,
using either a specified identity or a countertop reference. Similar
colors and category labels make branding and appearance important.
Targets are also placed in unexpected locations, such as Corn Flakes in
the fruit cabinet, to discourage shortcuts based on cabinet categories.

\item \textbf{Manipulation (3 tasks).}
Select where bread, a mug, or a bowl belongs using remembered cabinet
contents. For example, bread belongs with the pantry's dry goods even
though no cabinet already contains bread. The robot selects a destination
without transporting the object.

\item \textbf{Memory queries (4 tasks).}
Recall previously observed contents and spatial relationships, such as
what was stored with the mayonnaise or where the croissant near the
pineapple was located. These questions can be answered from existing
memory without further exploration.
\end{itemize}

\begin{figure}[H]
\centering
\includegraphics[width=0.95\linewidth]{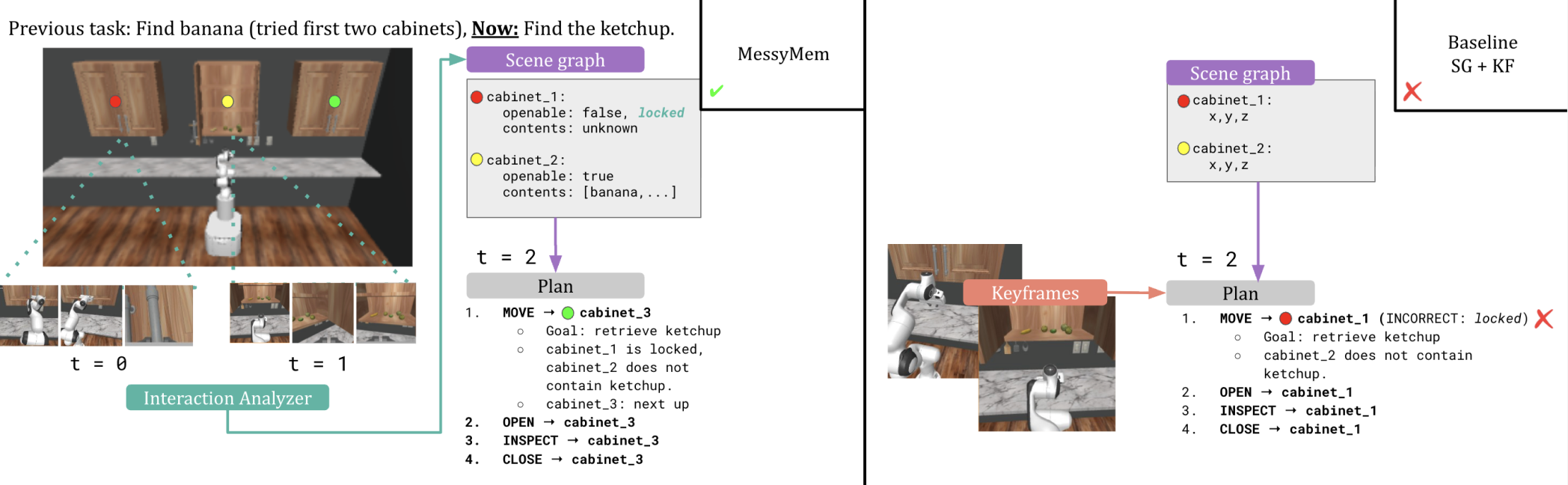}
\caption{
\textbf{Locked-and-Unlocked Cabinets.} This scenario highlights the role of the interaction analyzer. By reasoning over the interaction collage, \ApproachName{} correctly determines the first cabinet as locked and stores this in the scene graph. On the next task, the robot uses this memory, together with the remembered contents of the second cabinet, to skip both and move to the remaining cabinet.
}
\label{fig:sim-scenarios-locked-unlocked}
\end{figure}

\paragraph{Locked-and-Unlocked Cabinets.}
Three cabinets: \texttt{cab\_1} (far-left) is locked and empty, \texttt{cab\_2} (middle) holds the banana among fruit distractors, and \texttt{cab\_3} (far-right) holds the ketchup but is never observed during the first instruction. Spawning in front of \texttt{cab\_1}, the robot is first instructed \emph{``Look into upper cabinets and find the banana.''}. Its first attempt to open \texttt{cab\_1} fails (locked), after which it finds the banana in \texttt{cab\_2}, recording that \texttt{cab\_1} is locked and \texttt{cab\_2} lacks ketchup. Returned to the first cabinet, it is then instructed \emph{``Find the ketchup.''} Success requires navigating to the unexplored \texttt{cab\_3} while skipping the locked \texttt{cab\_1} and the ketchup-free \texttt{cab\_2}; re-attempting the locked cabinet is considered a failure. Figure \ref{fig:sim-scenarios-locked-unlocked} compares \ApproachName{} against \textbf{SG+KF}, showing the utility of the interaction analyzer. By reasoning about its first failed opening, \ApproachName{} records the left cabinet as locked and uses this memory on the next task when finding a new item. In contrast, \textbf{SG+KF} attempts to re-open the first drawer, failing the task.

\begin{figure}[H]
\centering
\includegraphics[width=0.95\linewidth]{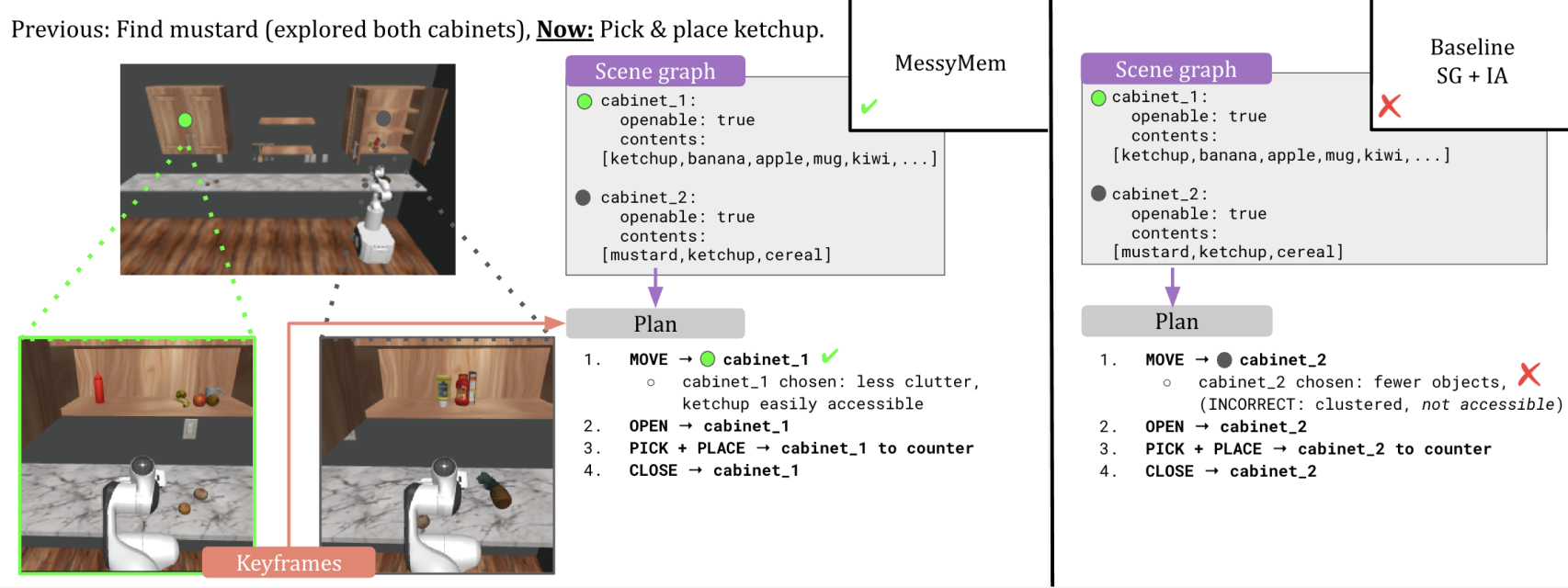}
\caption{
\textbf{Clutter-Aware Pick.} This scenario demonstrates the importance of keyframe memory. The robot has to look back at prior visual observations to compare fine-grained details such as clutter, free space, and object layout. The planner is able to query the relevant keyframes to determine picking from which cabinet is more accessible and less likely to disturb nearby objects.
}
\label{fig:sim-scenarios-executable-pick}
\end{figure}

\paragraph{Clutter-Aware Pick.}
Given two cabinets, both of which contain a ketchup bottle, but in \texttt{cab\_1} it sits alone with clear side access (five other items packed to one side), whereas in \texttt{cab\_2} it is wedged among two tightly-clustered items. The cluttered cabinet holds \emph{fewer} items overall, so item counts are anti-correlated with accessibility. The robot is first instructed \emph{``Look in the upper cabinets and find the mustard.''} (mustard is in \texttt{cab\_2}), opening both cabinets and seeding memory with interior keyframes. It is then instructed \emph{``Pick up the ketchup from a cabinet and place it on the counter.''}, ending in front of the already-open \texttt{cab\_2}. By choosing based solely on proximity, the robot will attempt the wrong, cluttered cabinet. A scene-graph only (\textbf{SG}) method cannot confidently determine the correct cabinet as it only records the broad category of "ketchup". Success requires comparing the two interiors from the stored keyframes and picking from the less-cluttered \texttt{cab\_1}, testing whether keyframe memory supplies spatial information beyond semantic labels and positions. Figure \ref{fig:sim-scenarios-executable-pick} compares \ApproachName{} against \textbf{SG+IA}, highlighting the impact of the keyframe memory. In this scenario, when needing to pick the ketchup, \ApproachName{} enables querying relevant keyframes for the VLM to decide which cabinet is more accessible and less cluttered for picking. In contrast, \textbf{SG+IA} solely relies on how many objects there are in a cabinet, which is an insufficient signal, and therefore picks up the ketchup that is cluttered amongst other condiments.

\subsection{Baselines}
\label{app:baselines}

All methods use the same environment layouts, perception front-end, primitive
skills, planner model, and task-success checks unless otherwise noted. We
compare against both internal ablations of \ApproachName{} and external memory
representations.

\paragraph{Internal ablations.}
Our ablations selectively remove components of \ApproachName{} while keeping
the remaining system fixed.

\begin{itemize}
    \item \textbf{\ApproachName{} (SG+IA+KF).}
    Our full method combines a persistent 3D scene graph, interaction-analyzer
    updates, and linked keyframe memory.

    \item \textbf{SG+IA.}
    Scene graph memory with interaction-derived updates, but without keyframes.
    This tests whether structured interaction outcomes alone are sufficient
    without access to fine-grained visual evidence.

    \item \textbf{SG+KF.}
    Scene graph memory with linked keyframes, but without interaction-analyzer
    updates. This tests whether visual recall can compensate for missing
    interaction-derived properties and outcomes.

    \item \textbf{SG.}
    A persistent scene graph containing object labels and positions, but no
    interaction-derived updates or keyframe memory. This tests the value of a
    spatially grounded memory without either interaction semantics or visual
    recall.
\end{itemize}

\paragraph{RoboEXP-ACSG.}
We adapt RoboEXP's Action-Conditioned Scene Graph (ACSG)~\cite{jiang2024roboexp}
from its released implementation. RoboEXP is designed for interactive
exploration, where actions reveal hidden scene structure that is incorporated
into a relational graph containing object identities, geometry, and coarse
relations such as \texttt{inside}, \texttt{on}, and \texttt{under}. We use the
ACSG as a drop-in replacement for \ApproachName{}'s memory while keeping
perception, planning, primitives, and evaluation fixed. Rather than running
RoboEXP's exploration policy end-to-end, we adapt its graph updates to our
shared stack and allow the graph to persist across sequential tasks and be
queried by the same closed-loop planner as \ApproachName{}, strengthening the
baseline beyond its original exploration-focused setting. Unlike
\ApproachName{}, RoboEXP-ACSG does not retain linked visual keyframes or richer
interaction-derived properties and outcomes, such as whether a fixture was
discovered to be locked.

\paragraph{MemER-style.}
We implement a MemER-style visual memory following the high-level memory
selection procedure of~\citet{sridhar2025memer}. We use the term
\emph{MemER-style} because we reproduce its memory mechanism without
fine-tuning or reproducing its original low-level policy. At each update, a
VLM identifies temporally informative frames; nearby selections are clustered
along the time axis, and representative frames are retained in a bounded
visual memory. We follow the original integer hyperparameters with a memory
length of 8, recent window of 8, and temporal merge distance of 5.

MemER was originally developed in a tabletop manipulation setting, where the
relevant workspace remains largely within view. Its memory therefore consists
of selected past images without a persistent spatial index associating those
images with navigable scene entities. In our mobile-manipulation setting,
objects and fixtures frequently leave the robot's current view, so generated
subtasks must be grounded from the current observation alone. We use the same
VLM and primitive execution stack as the other methods so that differences
primarily reflect the memory representation rather than the underlying model
or controller.

\paragraph{MemER+Fixtures.}
We additionally evaluate \textbf{MemER+Fixtures}, which uses the same visual
memory mechanism as MemER-style but additionally provides the planner with the
identities and world positions of the kitchen fixtures. This gives the method
an explicit spatial vocabulary for referring to and navigating toward
out-of-view destinations while leaving its visual memory selection unchanged.
The comparison between MemER-style and MemER+Fixtures measures how
much explicit spatial grounding helps when adapting their method to mobile manipulation.

\section{Real-World Scenario Details}
\label{app:real-eval}

\subsection{Task Descriptions}
\label{app:real-scenarios}
\paragraph{Office Drawer Search.}
The robot searches three office drawer units where the left drawer is locked, the middle and right are unlocked. The scissors are in the right drawer (with a green cup on top), a banana is in the middle drawer, and an orange cup labeled ``John'' sits atop the locked left drawer; before the final instruction, a gaming controller is placed in the right drawer. The robot receives three sequential instructions: \emph{``Find me scissors.''}, then \emph{``Move to John's cup.''}, then \emph{``Since you finished your last task, my friend just placed a gaming controller into one of the unlocked drawers. Help me find it.''} Across these it must remember which drawer failed to open (locked), what contents were already observed, and which locations remain plausible for the next target.

\paragraph{Sock Pairing.}
Three fixtures are arranged left-to-right, two drawers and a chair, with socks in each drawer and one lone sock placed on the chair. The lone sock's distinguishing attribute (color or pattern) and which drawer holds its match are swapped and randomized between trials, so the answer cannot be memorized and must be inferred from observation. The robot receives three sequential instructions: \emph{``Move to each drawer unit, open it, and peek inside to observe the contents.''}, then \emph{``Move to the chair.''}, then \emph{``Pick up the lone sock from the chair and place it into the drawer with its matching pair.''} Success requires visual memory. The planner must recall which drawer contains the sock matching the lone sock, a distinction of color or pattern rather than object category alone.

\end{document}